\PassOptionsToPackage{T2A}{fontenc}
\documentclass[onecolumn]{cohere}
\renewcommand{\fasymbol}{\textcolor{ayadsymbol}{$\vardiamond$}}

\usepackage{lipsum}
\usepackage{pdfpages}
\usepackage{colortbl}
\usepackage{tabulary}
\usepackage{longtable}
\usepackage{array}
\usepackage{placeins}
\usepackage{tablefootnote}
\usepackage{tcolorbox}
\usepackage{wrapfig}
\usepackage{breqn} 

\usepackage{pifont}
\definecolor{deeppurple}{HTML}{9e02f7}
\definecolor{forestgreen}{HTML}{2e7d43}
\usepackage{multirow}
\usepackage{tabularx}

\usepackage[utf8]{inputenc}

\usepackage{arydshln}

\usepackage{xspace} 
\usepackage{makecell} 
\usepackage{multirow}

\usepackage[framemethod=TikZ]{mdframed}
\usepackage{tcolorbox}
\usepackage{multicol}

\colorlet{LightGreen}{green!20}
\colorlet{LightRed}{red!20}
\colorlet{LightGrey}{black!20}
\tcbset{on line,
        boxsep=1.5pt,left=0pt,right=0pt,top=0pt,bottom=0pt,
        colframe=white
        }

\newtcolorbox{mybox}[2][]{
  colback=white, 
  colframe=lightblue,
  fonttitle=\bfseries,
  coltitle=black,  
  title=#2, 
  #1
}

\definecolor{ayad}{RGB}{148, 156, 229} 
\definecolor{ayadsymbol}{RGB}{76, 110, 230} 
\definecolor{lightblue}{RGB}{211, 227, 252} 
\definecolor{bgblue}{RGB}{247, 250, 255} 
\newcommand*\colourcheck[1]{%
  \expandafter\newcommand\csname #1check\endcsname{\textcolor{#1}{\ding{52}}}%
}
\newcommand*\colourcross[1]{%
  \expandafter\newcommand\csname #1cross\endcsname{\textcolor{#1}{\ding{55}}}%
}
\DeclareSymbolFont{extraup}{U}{zavm}{m}{n}
\DeclareMathSymbol{\vardiamond}{\mathalpha}{extraup}{87}
\definecolor{ayadsymbol}{RGB}{76, 110, 230} 

\colourcheck{blue}
\colourcheck{green}
\colourcheck{red}

\colourcross{blue}
\colourcross{green}
\colourcross{red}
\usepackage{nicematrix}
\usepackage{comment}
\usepackage{amssymb}

\setcitestyle{number,square}

\usepackage{epigraph}
\usepackage{comment}

\title{Multilingual Arbitrage: Optimizing Data Pools to Accelerate Multilingual Progress}

\multiauthors
\author{
    name={\Large Ayomide Odumakinde \fa},
    affiliation={1}
}

\author{
    name={\Large Daniel D'souza \fa},
    affiliation={1},
}
\author{
    name={\Large Pat Verga},
    affiliation={2},
}
\author{
    name={\Large Beyza Ermis \fa},
    affiliation={1},
}
\author{
    name={\Large Sara Hooker},
    affiliation={1},
}

\affiliations{
    \large \item[1] Cohere For AI 
    \item[2] Cohere
}

\corresponding[*]{\centering Beyza Ermis, Sara Hooker, Ayomide Odumakinde \{\url{beyza}, \url{sarahooker}, \url{ayomideodumakinde}\}\url{@cohere.com}}

\date{\today}
\abstract{ 
The use of synthetic data has played a critical role in recent state-of-art breakthroughs. However, overly relying on a single \textit{oracle} teacher model to generate data has been shown to lead to model collapse and invite propagation of biases. These limitations are particularly evident in multilingual settings, where the absence of a universally effective teacher model that excels across all languages presents significant challenges. In this work, we address these extreme difference by introducing ``\textit{multilingual arbitrage}'', which capitalizes on performance variations between multiple models for a given language. To do so, we strategically route samples through a diverse pool of models, each with unique strengths in different languages. Across exhaustive experiments on state-of-art models, our work suggests that \textit{arbitrage} techniques allow for spectacular gains in performance that far outperform relying on a single teacher. In particular, compared to the best single teacher, we observe gains of up to 56.5\% improvement in win rates averaged across all languages when switching to multilingual arbitrage. We observe the most significant gains for the least resourced languages in our pool.}

\begin{document}

\epigraph{{\fontfamily{cmr}\selectfont Two heads are better than one, not because either is infallible, but because they are unlikely to go wrong in the same direction.}}{\textit{C.S. Lewis}}

\section{Introduction}
\label{sec:intro}

Throughout our lives, we are guided by many teachers, each contributing distinct insights and expertise to our personal and professional growth. For specialized skills, such as becoming a doctor or mastering culinary arts, we seek out experts who provide targeted guidance. In contrast, synthetic data generation often relies on a single teacher model to impart knowledge to a student model. This approach can lead to the passive transfer of both the strengths and limitations inherent in the teacher model, as highlighted in various studies~\citep{shumailov2023curse,magister2023teaching,shimabucoro2024llmseellmdo,gerstgrasser2024model}. Moreover, it assumes that a single model can effectively teach all relevant skills, which may not always be the case. 

\begin{figure*}[ht!]
    \centering
         \includegraphics[width=1.0\textwidth]{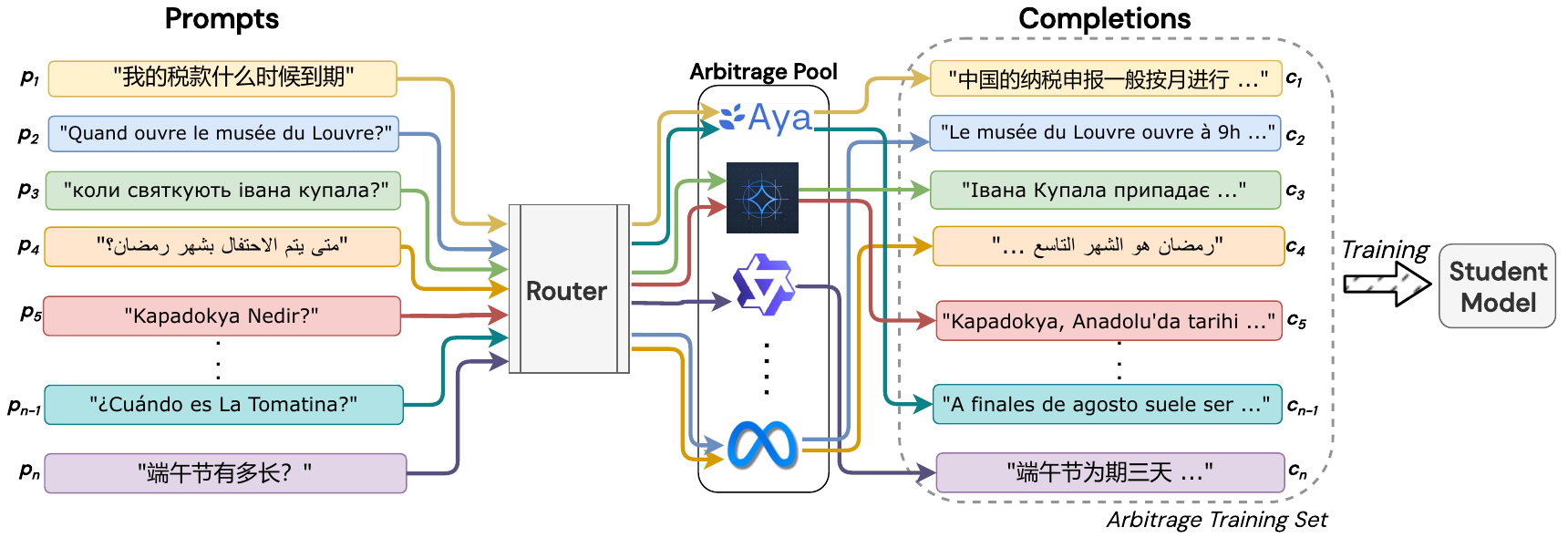}
         \caption{\textbf{Overview of \emph{Multilingual Arbitrage}.} Instead of relying on a single ``oracle'' teacher, multilingual arbitrage re-frames the distillation problem as learning how to optimize sampling for a desired part of the data distribution from an ensemble of teachers.}
         \label{fig:multilingual-arbitrage}
\end{figure*}

The limitations of the single oracle approach become particularly pronounced in multilingual settings, where high-performing large language models (LLMs) are often trained predominantly on a few data-rich languages~\citep{singh2024aya,joshi2020state,fan2021beyond}. This diverse landscape of multilingual model development has resulted in a variety of models: large-scale models that support multiple languages~\citep{xue2020mt5,scao2022bloom,shliazhko2022mgpt,li2023bactrianx,ustun2024aya}, frontier models with some multilingual capabilities that are not specifically optimized~\citep{armengol2021multilingual,chowdhery2022palm,zhang2022opt,team2024gemma}, and models focused on regional language families~\citep{adelani2021masakhaner,mirzakhalov2021large,sea_lion_2024,cahyawijaya2022nusacrowd}. As a result, it is often unclear how to determine which model to use to maximize performance for a given language. Relying on a single model can also further amplify disparities in treatments between languages, as models may perform well on some language but not have coverage for others. Performance tends to be critical for the quality of synthetic data, which can enable further progress in those languages by making data more ubiquitous over time~\citep{alaa2022faithful,gao2023selfguidednoisefreedatageneration,bukharin2023data,zhang2023auto,li2024synthetic,zhang2024instruction}. 

\textbf{In this work, we take a wider view of synthetic data generation.} Rather than perceiving model distillation as the transfer of data from a single oracle to a student, we re-frame the problem within this heterogeneous landscape as learning how to optimize sampling for a desired part of the data distribution from an ensemble of teachers. Multilingual settings offer a valuable case study for this exploration due to the clearer boundaries between languages compared to tasks. We expect our \textit{arbitrage techniques} to lead to improvements across any setting where it would be unusual for a single model to be state-of-art at all tasks. 

We introduce the concept of \textit{multilingual arbitrage}, which leverages differences in performance for a language between multiple models. For each language, we utilize a pool of models as potential teachers. We evaluate different methods for sampling strategically by routing to different models. We then use this optimized distribution to instruction fine-tune a new multilingual model, with the goal of exceeding the performance achieved by relying on a single multilingual model across all languages. This also allows us to ask – \textit{can you surpass individual model performance by sampling parts of the distribution strategically from a pool of models?}

We conducted exhaustive experiments across 15 languages using 9 state-of-the-art multilingual models to evaluate our method. Our extensive evaluations included LLM-as-an-evaluator win rates, discriminative tasks, and textual characteristics. Our key findings and contributions are as follows:

\begin{itemize}
\item \textbf{We introduce the concept of ``multilingual arbitrage'' which significantly outperforms traditional single teacher distillation.} Across a range of experimental settings, we demonstrate that our arbitrage methods significantly outperform single teacher models. Specifically, our reward-based routing technique achieved an average improvement of 56.5\% in generative win-rates and a 28.1\% improvement compared to the best single teacher model. On discriminative tasks, the reward-based routing led to improvements of up to 3.27\%, with an average improvement of 1.91\% over the base model, while single teacher models showed an average improvement of only 0.98\%.

\item \textbf{Not all arbitrage techniques are equal.} 
We evaluate the performance of various arbitrage techniques. Against a lower bound baseline of random routing, we observe that reward-based routing, fixed routing with a predefined set of expert teachers, and learned routing improved performance by 119.5\%, 76.8\%, and 40.6\%, respectively. Although reward-based filtering, which is more resource-intensive, proved to be the most effective, our results indicate that the far more efficient reward-guided learned routing can achieve impressive performance gains without the need to generate all completions from each model.

\item \textbf{Arbitrage improves or maintains textual characteristics.}
Lastly, we analyze the textual characteristics of student model generations by calculating various statistics and readability scores. This analysis highlights the effects of instruction fine-tuning (IFT) with multilingual arbitrage on text verbosity, readability, and lexical diversity of generations. 
Our findings indicate that reward-based routing results in a 14.1\% increase in the number of tokens in the generated text, while learned routing leads to a 68.4\% increase compared to both single-teacher generations (averaged across all single teachers) and random routing.
Additionally, we observe an increase in lexical diversity scores: reward-based routing improves scores by 6\% and learned routing by 4.2\% compared to single teachers, and by 13.4\% and 11.5\% compared to random routing, respectively.
\end{itemize}

\section{Methodology\label{sec:methods}}

Our primary goal is to train a high-performing multilingual student model $S$. Given a set of input prompts $P = \{p_i\}_{i=1}^N$, we generate a corresponding set of completions $C = \{c_i\}_{i=1}^N$ using a pool of potential teacher models $\mathcal{T} = \{T_j\}_{j=1}^M$. These prompt-completion pairs $(p_i, c_i)$ will then be used to fine-tune $S$. For each prompt $p_i \in P$, we aim to identify the specific teacher model $T_j \in \mathcal{T}$ that produces the highest quality completion $c_i$. 

We consider that each teacher model $T_j$ may not perform uniformly across all regions of interest $R$ in the data distribution. Therefore, we aim to minimize the empirical error $E[P_j(R)]$, where $P_j(R)$ represents the performance of teacher model $T_j$ in region $R$, over the broader distribution $D$. This ensures robustness and generalization beyond the i.i.d. training sample $D_{\text{iid}}$. This approach allows us to select the most suitable teacher model for each prompt, optimizing the training of our student model $S$. We note that this amounts to optimization in the data space and allows for \textit{on-the-fly} creation of dataset properties to minimize sensitivity to distribution drift. 

\begin{figure*}[ht!]
    \centering
         \includegraphics[width=1.0\textwidth]{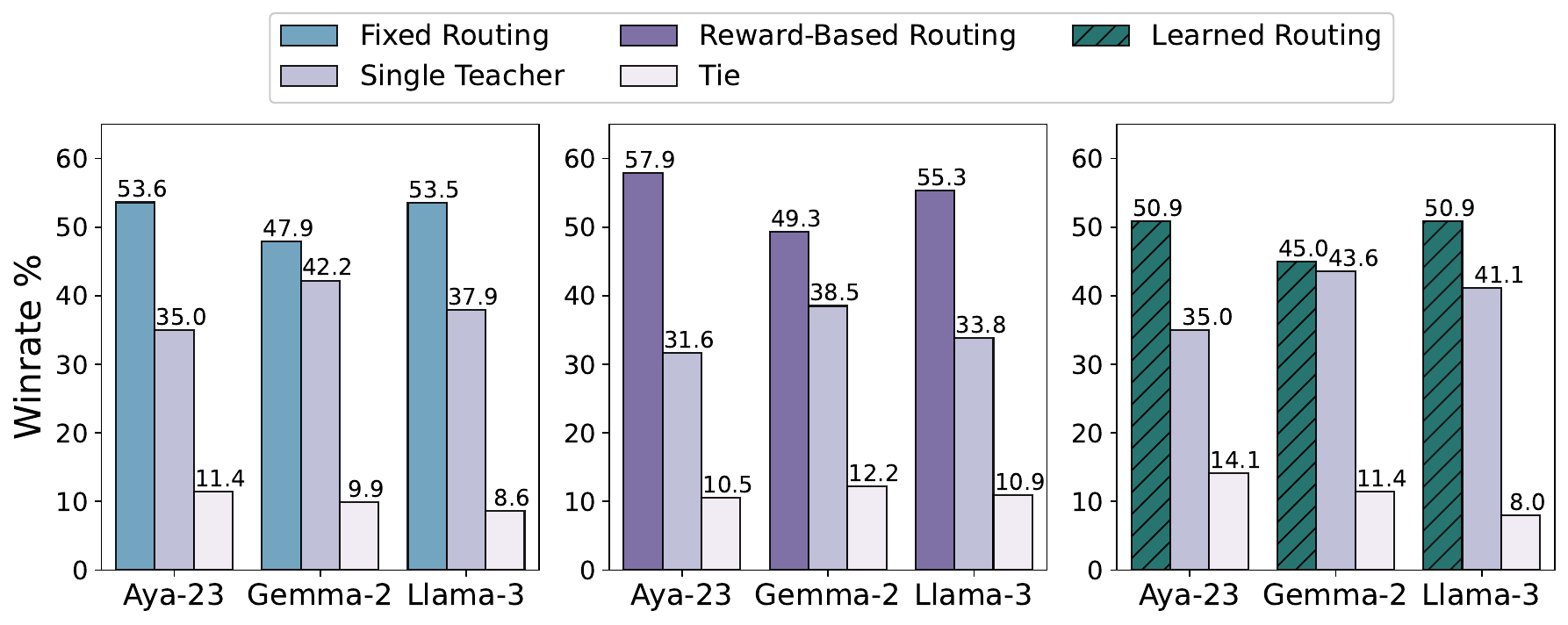}
         \caption{\textbf{Win rate \% comparison of Fixed Routing, Reward-Based Routing and Learned Routing against multiple Single Teacher Models.} The x-axis represents the single teacher used to generate the synthetic data for training. We observe that all multilingual arbitrage strategies \textit{outperform all} single teachers with the largest gains observed for reward-based routing. All values are percentages and aggregated across 7 languages: \emph{Arabic, Chinese, English, French, German, Turkish, and Ukrainian}.}
         \label{fig:winrates-single-teachers}
\end{figure*}

\subsection{Routing Methods \label{sec:routing}}

The crux of the problem of multilingual arbitrage is: \textit{how do you route prompts to the most calibrated teacher model for each prompt?} We exhaustively benchmark different routing strategies which we introduce briefly below: 

\noindent\textbf{Fixed Routing.}
In practice, one might choose a fixed model, such as $T_2$, to process all input prompts in $P$. This can be reasonable if $T_2$ demonstrates significantly better overall performance for a majority of the prompts. In the multilingual case, this setting is one in which we can pre-determine the best model for each language based on their known strengths, enabling us to use a fixed routing strategy for each prompt deterministically by choosing the appropriate teacher model according to the prompt's language. However, in real-world settings it is not always possible to know what models are relatively strong at different languages in advance.

\noindent\textbf{Reward-based Routing.}
Next we consider the more realistic setting which assumes that we cannot pre-determine a fixed routing strategy. Instead, we rely on a reward model for routing. For each $p_i$ we generate a completion from each of the teacher models in $\mathcal{T}$ and then select $c_i$ to be the completion with the highest score given by some ranking method. In our case, we use a proprietary reward model (Cohere May 2024) which is competitive with top-scoring state-of-the-art reward models on the RewardBench Leaderboard \citep{lambert2024rewardbench}\footnote{\url{https://huggingface.co/spaces/allenai/reward-bench}}. We intentionally use a separate reward model for routing from the model that we use for our LLM-as-a-judge evaluation (GPT-4-Turbo7\footnote{\url{https://platform.openai.com/docs/models/gpt-4-turbo-and-gpt-4}}) given the known biases incurred by using the same model for both~\citep{bansal2023peering,verga2024replacing,shimabucoro2024llmseellmdo}.

\begin{table*}[t]
\small
\centering
\setlength{\tabcolsep}{4pt}
\begin{tabular}{@{}lccccc@{}}
\toprule
& Fixed & Reward Score & Learned \\ \midrule
Works with Unknown Teachers  & \redcross &  \bluecheck & \bluecheck \\ [0.1cm]
All models are considered for each prompt & \redcross & \bluecheck & \bluecheck   \\ [0.1cm]
Efficient Routing & \bluecheck & \redcross & \bluecheck   \\ [0.1cm]
New models can be added on-the-fly &  \redcross & \bluecheck & \redcross \\ 
\bottomrule
\end{tabular}
\caption{\small{\textbf{A comparison of different arbitrage techniques:} We compare the properties of the different proposed routing methods. While the reward-based routing is the most flexible approach, it comes at the cost of efficiency as compared to the learned router.}}
\label{tab:intro}
\end{table*}

\noindent\textbf{Learned-Routing.} 
The disadvantage of reward-based routing is that it requires generating a full set of $M$ completions for each prompt where $M=|\mathcal{T}|$. As a more efficient alternative, we explore the merits of a \textit{learned router} which instead trains a router model based on scores produced by the reward model which is proposed by~\citep{lu-etal-2024-routing}. 
In this method, the router model learns to predict the reward conditioned only on the prompt $p_i$, thereby determining the most suitable teacher model $T_j$ without the need to generate multiple completions based upon historical routing trends.
The router $R(p_i)$ is defined to select the teacher model $T_j$ that maximizes the expected reward for a given prompt $p_i$. Formally, for each $p_i \in P$, the selected model $T_j$ is given by:
$$
T_j = \arg\max_{T \in \mathcal{T}} R(p_i, T).
$$
This approach leverages the complementary strengths of the models in $\mathcal{T}$ and ensures that each prompt is routed to the model most likely to produce the highest quality completion. By integrating reward model ranking with query routing, reward-guided Learned-Routing enhances the efficiency of the LLM ensemble, reducing computational overhead while ensuring effective training of the student model $S$.

To train our learned-routing model, we collect a training dataset of diverse prompts and then generate completions from each of the candidate models in the teacher pool. Given a prompt from our training set, we obtain a scalar reward for each candidate model generation as in the following:
\begin{equation}
    \mathbf{r_i} = \left\{RM(p_i, T_j(p_i)) \right\}_{j=1}^{|\mathcal{T}|}, \quad i = 1, \dots, N
\end{equation}
where $\mathbf{r_i} \in \mathcal{R}^{|\mathcal{T}|}$.  We then train our router $R$ on the training data with Kullback-Leibler (KL) divergence as the loss function: 
\begin{equation}
    \mathcal{L}(p_i, \mathbf{r}_i) = \text{KL}(R(p_i), \text{softmax}(\mathbf{r}_i)).
\end{equation}
This approach improves the quality of synthetic data while maintaining computational efficiency during inference, introducing only minimal overhead compared to traditional reward model ranking methods, which is training the router model. However, this overhead is well compensated during inference because learned routing only generates samples from the routed model, rather than from each model in the pool. As a result, the generation cost is reduced to $1/M$, where $M$ is the number of models in the pool.

\section{Experimental Setup \label{sec:experiment_setup}}

\subsection{Baselines}
\label{subsec:single_teacher}

To evaluate the effectiveness of \textit{multilingual arbitrage}, we compare our method against several baseline methods. Below, we provide a brief overview of the experimental details for each baseline:

\textbf{Single Teachers.} 
This is the most widely adopted framework for incorporating synthetic data into training. In this paradigm a student model is trained on the generations produced from a single teacher model. This setup allows us to explore the question: \textbf{Is multilingual arbitrage more effective than a single \textit{``oracle''} teacher?}

We choose single teacher models based on architecture, size, base model type, and the extent of coverage of languages. The selected models, ranging from 7B to 9B parameters, are widely used: Aya 23~\citep{aryabumi2024aya}, Llama 3~\citep{dubey2024llama}, and Gemma 2~\citep{team2024gemma}. We include more details about each model in Appendix Section \ref{app:teacher_pool}. We note that both Llama 3 and Gemma 2 do not explicitly claim multilingual support, however, in practice both are more frequently used by multilingual users compared to models explicitly designed for multiple languages, such as mT0~\citep{muennighoff2022crosslingual} and BLOOMZ~\citep{dac2023okapi}.

\textbf{Random Routing.} Next, we consider a router that \textbf{randomly} assigns each prompt $p_i \in P$ to teacher model $T_j \in \mathcal{T}$, without considering the language or any other specific characteristics of the prompt. This allows us to investigate: \textbf{Is multilingual arbitrage better than a random selection as to what model is best for a given distribution of interest?}

\textbf{Translation.} Lastly, our translation baseline addresses whether strategic sampling outperforms simply translating the generations of a single English model into multiple languages. We aim to determine: \textbf{Does generating synthetic data in the target language outperform translating the best English only data?} 

We generate completions for our English training prompts using our most capable English teacher model, Llama 3. We then translate each of the prompts and completions to the seven languages included in our router experiments. 

\subsection{Routing Teacher Pools}

\textbf{Fixed Router Model Pool.}\label{subsec:fixed_router} In our fixed router experiments, we assume prior knowledge about which models perform best for specific languages. To create this setting, we train several geo-cluster models on 15 languages, each specializing in a subset of languages corresponding to the language families detailed in Table \ref{table:clusters}.
Training in this way allows the models to leverage geographic and linguistic similarities within a language cluster \citep{kohli2023buildingllama2finetunedllmodia, kew2023turningenglishcentricllmspolyglots, tejaswi2024exploringdesignchoicesbuilding}. 
Each geo-cluster surpasses the single teacher model prior to student model training, achieving an average win rate gain of 14.9\% relative to single teachers. See Appendix \ref{app:geocluster_training} for additional training and win rate evaluation details.

\begin{table}[h]
\centering
\begin{tabular}{ll}
\toprule
\textbf{Language Cluster} & \textbf{Languages} \\
\midrule
\textsc{Germanic} & German, Dutch \\
\textsc{Slavic} & Czech, Russian, Ukrainian, Polish \\
\textsc{Romance} & French, Portuguese, Spanish, Italian, Romanian \\
\textsc{East-Asian} & Korean, Japanese, Chinese, Turkish \\
\bottomrule
\end{tabular}
\caption{\textbf{Language composition of Geo-clusters:} To evaluate fixed routing, we control apriori for the strength of a model on each language in our pool by training Geo-cluster models which are specialized on different groups of languages.}
\label{table:clusters}
\end{table}

\textbf{Reward-based and Learned Routing.} These methods aims to demonstrate the effectiveness of routing in a varied pool of models with unknown multilingual performance. Hence, we consider a diverse pool that includes all single teacher models (\ref{subsec:single_teacher}), geo-cluster models (\ref{subsec:fixed_router}) and monolingual models in Chinese (Qwen2-7B-instruct \citep{yang2024qwen2technicalreport}) and Turkish (Turkish-Llama-8b-Instruct-v0.1 \footnote{\url{https://huggingface.co/ytu-ce-cosmos/Turkish-Llama-8b-Instruct-v0.1}}) which are specifically designed to support individual languages. We include more details about the monolingual models in Appendix \ref{app:teacher_pool}. This variety, ranging from massively multilingual to geo-cluster and monolingual models, helps us analyze which types of models are most utilized by different routing techniques.

\textbf{Learned Routing} To train our learned router, we fine-tune the Gemma2-2B\citep{team2024gemma} model, selected for its compact size, strong performance, and multilingual capabilities. To further improve training efficiency, we also evaluate a smaller mT5-base~\citep{xue2020mt5} variant with 580M parameters. 
Comparative results for these models are presented in Appendix~\ref{app:router-model-details}.
Our learned router models were trained using prompts from Dolly-15k which were translated using NLLB-3.3B~\citep{nllb2022no} into the seven languages covered by our routing experiments, and resulting in 60,419 prompts in total.

\subsection{Student Model}
\label{sec:student_models}

We selected the Aya 23 8B model~\citep{aryabumi2024aya} as our student model due to its strong multilingual capabilities which are state-of-the-art (SOTA) for its size class. For each arbitrage technique, we generate synthetic data in seven diverse languages: \emph{Arabic, Chinese, English, French, German, Turkish}, and \emph{Ukrainian}. These languages, representing different language families, are selected to ensure a comprehensive evaluation across various linguistic contexts. Detailed information is provided in Table~\ref{table:language_lineage} in Appendix~\ref{app:language_fam}.

\textbf{Training Details.} Our student models are trained from a collection of prompts derived from \textit{UltraFeedback Binarized Dataset}~\citep{tunstall2023zephyr}, an English preference dataset of 61,135 pairs. We randomly sampled 10,000 prompts and translated them into the seven target languages using the NLLB-3.3B model, resulting in a total of 70,000 prompts. Completions for each prompt were then generated by the assigned teacher model. We then instruction fine-tune each student model on these 70,000 data points selected through multilingual arbitrage. 
The training employed a cosine learning rate schedule with a warm-up phase, using a batch size of 64 and an evaluation batch size of 128. The peak learning rate was set at $2.5 \times 10^{-5}$, achieved through 128 warm-up steps starting from a learning rate of $0.0$, and then decayed back to $0.0$. 

\subsection{Evaluations \label{sec:evals}}

\textbf{Open-ended Generation Win rates.} Beyond traditional NLP tasks, we aim to evaluate the open-ended generation capabilities of the student models, focusing on their ability to produce unstructured and long-form responses. For this evaluation, we use GPT-4 as an LLM-judge to measure pairwise win rates between two model generations.
We evaluate on the target language subset of the Multilingual Dolly-200 Eval dataset~\citep{singh2024aya,ustun2024aya}. This 200 instance evaluation dataset is a held-out curated sample from the Dolly-15k dataset~\citep{costa2022no}.
These prompts are open-ended and capture general-purpose non-code use cases. Hence, evaluation using this dataset is a valuable proxy for how multilingual arbitrage impacts more fluid and often open-ended asks. 

\textbf{Discriminative Tasks.} To evaluate our models on completely unseen tasks, we follow~\citet{muennighoff2022crosslingual} and use XNLI~\citep{conneau2018xnli}, XCOPA~\citep{ponti2020xcopa}, and XStoryCloze~\citep{lin2021fewshot} datasets targeting natural language inference, commonsense reasoning and sentence completion respectively. 
These unseen tasks are crucial for evaluating the effectiveness of IFT in improving a model's reasoning and comprehension capabilities as they test the model's ability to discriminate between different possible interpretations or outcomes. For all unseen tasks, we report zero-shot performance.

\section{Results and Discussion}

\subsection{Multilingual Arbitrage Performance} 

\begin{tcolorbox}[colframe=gray!90, colback=gray!20, title=Section Findings, fonttitle=\bfseries, boxrule=0.5pt, left=5pt, right=5pt, top=5pt, bottom=5pt, boxsep=5pt]
\begin{itemize}[left=0pt, topsep=2pt, itemsep=3pt, parsep=0pt]
    \item The multilingual arbitrage methods consistently outperformed the random routing baseline, achieving an average win rate of 51.8\% with a notable improvement of 78.9\% over random routing.
    \item Arbitrage strategies demonstrate substantial improvements over single teacher models. Reward-based routing achieves the highest average win rate gains at 56.5\%, while learned routing provides a notable absolute gain of 25.6\% over single teacher models and is significantly more efficient - requiring only one generation per prompt, making it nine times more efficient than reward-based routing.
    \item On unseen discriminative tasks, arbitrage methods provided a larger average improvement of 1.95\% over the base student model compared to a 0.98\% improvement by single teachers. While fixed routing achieved the highest average gain of 2.50\%, all routing approaches effectively enhanced cross-lingual and commonsense reasoning capabilities, showcasing their overall superiority to single teacher models.
\end{itemize}
\end{tcolorbox}

\begin{figure}[htb]
    \centering
         \includegraphics[width=0.8\textwidth]{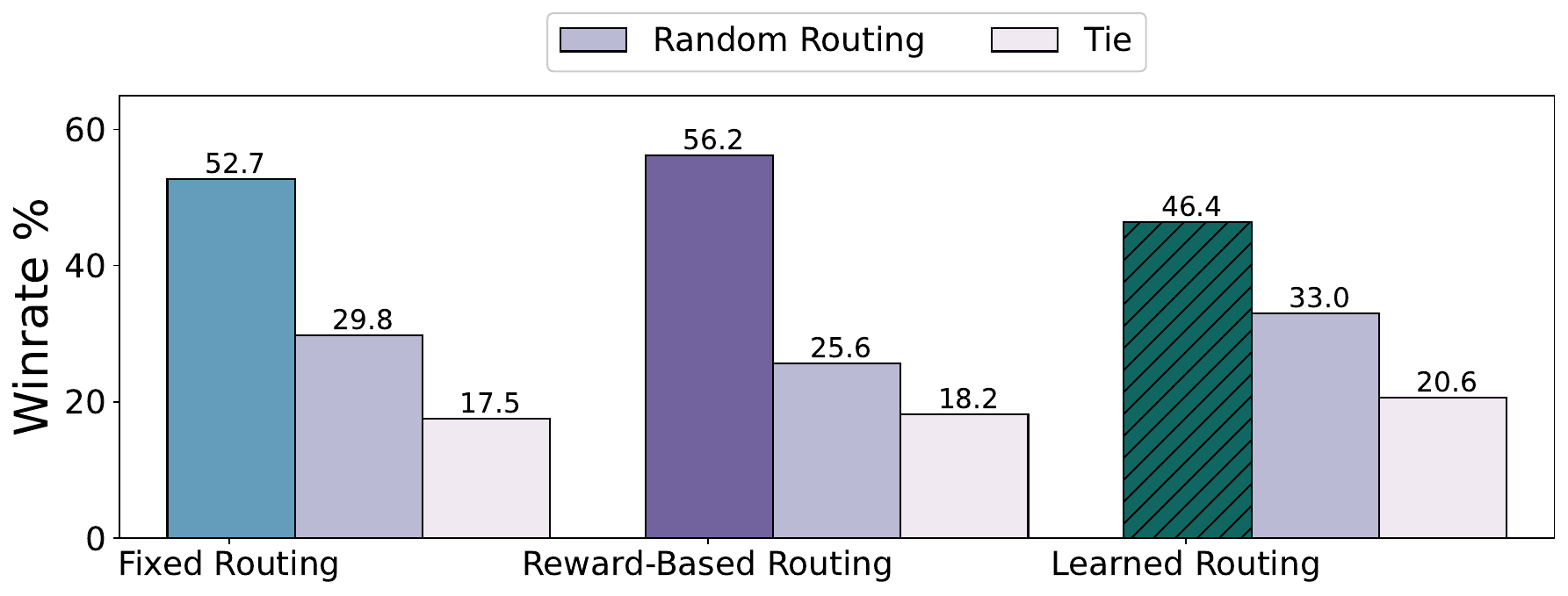}
         \caption{\textbf{Win rates \% of students trained with different routing strategies:} Comparison of router-trained students to those trained with random routing. The largest gains are observed for reward-based routing with a win-loss diff of 30.6\%. All values are percentages and aggregated over 7 languages.}
         \label{fig:winrates-random-routing}
\end{figure}

\textbf{Comparison against random routing.} 
Our random routing baseline serves as a crucial lower bound that any proposed arbitrage strategy should outperform. This baseline helps us evaluate: \textit{Is our multilingual arbitrage technique better than a random guess?} 
In Figure~\ref{fig:winrates-random-routing}, we compare the win rates of each of the different routing methods against the random routing baseline. We observe that all the multilingual arbitrage methods consistently outperformed the random baseline with average win rate of 51.8\% and a notable relative win rate improvement of 78.9\%.

\textbf{Comparison against single ``oracle'' teacher.}
In Figure \ref{fig:winrates-single-teachers}, we show win rates comparing our arbitrage routing strategies to single teacher models. Student models trained with data from these strategies significantly outperformed those using single teacher generations. Specifically, fixed routing achieves an average gain of 34.7\%, reward-based routing shows a 56.5\% improvement, and learned routing has a 25.6\% improvement in average over all single teachers. Notably, Gemma 2 was the best-performing single teacher, yet learned routing still achieved a 3.2\% gain over it.

\textbf{Win-rate Gains are largest for Reward-Based Routing.} 
We observed the largest improvements against single teachers with reward-based routing, achieving average gains of 56.5\%. However, reward-based routing is the least efficient arbitrage method because it requires running inference and generating completions with all models in the pool for each prompt. Although fixed routing and learned routing show some decrease in win-rates compared to reward-based routing, they are significantly more efficient during inference, as they only require inference from one model.
In our experiments with a pool of 9 models, reward-based routing requires generating and scoring 9 completions per prompt, while fixed and learned routing need only one generation per prompt. Although learned routing involves an additional call to the router per prompt, this router model is much smaller and more efficient than the teacher, making the call negligible compared to generating from all models in the pool. Notably, learned routing is the most flexible technique, being 9 times more efficient than reward-based routing in this setup and not needing prior knowledge of each model's merits, unlike fixed routing.

\textbf{Discriminative tasks.} Table~\ref{tab:discriminative-results} presents average scores for unseen discriminative tasks, reporting zero-shot performance. These tasks reveal similar gaps between the benefits of single teachers and arbitrage techniques. Single teachers provide an average improvement of 0.98\% over the base student model (Aya 23), while arbitrage techniques achieve a larger average improvement of 1.95\%.

\begin{table}[ht!]
    \centering
    \scalebox{0.9}{
    \begin{tabular}{lcccc}
        \toprule
         & \textbf{XCOPA} & \textbf{XNLI} & \textbf{XStoryCloze} & \textbf{Average} \\
        \midrule
        \noalign{\smallskip} 
        \textsc{AYA23} (Base Model) & 64.1 &  42.9 &  68.23 &   58.41 \\ \hline
        \noalign{\smallskip} 
        \noalign{\smallskip} 
        \textsc{Single Teachers} & 65.5  &  43.96  & 67.41  &  58.98 \tcbox[colback=LightGreen]{\textcolor{black}{$\uparrow 0.98$}} \\
        \noalign{\smallskip}         
        \textsc{Random Routing} & 65.9  & 44.01  & 67.25 
        &  59.05 \tcbox[colback=LightGreen]{\textcolor{black}{$\uparrow 1.09$}}\\
        \noalign{\smallskip}
        \midrule
        \textsc{Fixed Routing} & \textbf{67.4} & 43.89  & 68.33  
        & \textbf{59.87} \tcbox[colback=LightGreen]{\textcolor{black}{$\uparrow 2.50$}}\\
        \noalign{\smallskip}
        \textsc{Reward Based Routing} & 66.2 & \textbf{44.21}  &  68.20
        &  59.53 \tcbox[colback=LightGreen]{\textcolor{black}{$\uparrow 1.91$}} \\
        \noalign{\smallskip} 
        \textsc{Learned Router} & 65.8  & 43.62  & \textbf{68.36} 
        &  59.25 \tcbox[colback=LightGreen]{\textcolor{black}{$\uparrow 1.43$}} \\
        \noalign{\smallskip}
        \bottomrule
    \end{tabular}}
    \caption{\textbf{Performance of Student Models on held-out Discriminative Tasks:} XCOPA, XNLI, and XStoryCloze. All results are averaged over seven languages, highlighting the improvements or declines in performance compared to the base model AYA23. Single teacher results are the average of Aya23, Llama 3 and Gemma 2 single teachers. The `Average' column additionally includes the percent increase over the base model.}
    \label{tab:discriminative-results-short}
\end{table}

Overall, on discriminative tasks Fixed Routing model emerges as the most effective, with the highest average improvement of 2.50\% across all tasks, followed by reward based routing with 1.91\% improvement indicating their superior ability to enhance cross-lingual and commonsense reasoning capabilities in the evaluated student models. 
Notably, while fixed routing ranks first in discriminative tasks, it is second in win rate comparisons.
This discrepancy may stem from a noted tension between model performance on academic benchmarks and open-ended generations. Recent studies suggest that as performance on open-ended tasks improves, traditional academic task performance may decline~\citep{ustun2024aya, ouyang2022training, iyer2023optimlscalinglanguagemodel,muennighoff2022crosslingual}.
This occurs because supervised fine-tuning of large language models has increasingly been torn between objectives: improving traditional academic benchmarks and training LLMs to follow instructions, acquire conversational abilities, and be helpful and harmless \citep{aakanksha2024multilingualalignmentprismaligning}. See Table~\ref{tab:discriminative-results} in Appendix~\ref{app:discTasks} for comprehensive results.

\subsection{Language and Routing Analysis}
\label{sec:language_and_routing}

\begin{tcolorbox}[colframe=gray!90, colback=gray!20, title=Section Findings, fonttitle=\bfseries, boxrule=0.5pt, left=5pt, right=5pt, top=5pt, bottom=5pt, boxsep=5pt]
\begin{itemize}[left=0pt, topsep=2pt, itemsep=3pt, parsep=0pt]
    \item Medium-resource languages benefit more from routing strategies than high-resource languages. Specifically, reward-based routing achieves a 56.1\% gain over single teacher models for medium-resource languages, compared to a 35.7\% gain for high-resource languages. 
    \item Different models are favored for different languages, highlighting the strengths of arbitrage that leverages a diverse model pool. For example, Aya 23 is predominantly used for Ukrainian, Turkish, and Arabic, while Llama3 is favored for English prompts.
    \item Generating synthetic data directly in target language is more effective than translating English-only data. Even a single teacher model achieves a 48.9\% increase in win rate compared to translations made from English data generated by the top-performing English model.
\end{itemize}
\end{tcolorbox}

\begin{figure}[h!]
    \centering
        \includegraphics[width=0.9\textwidth]{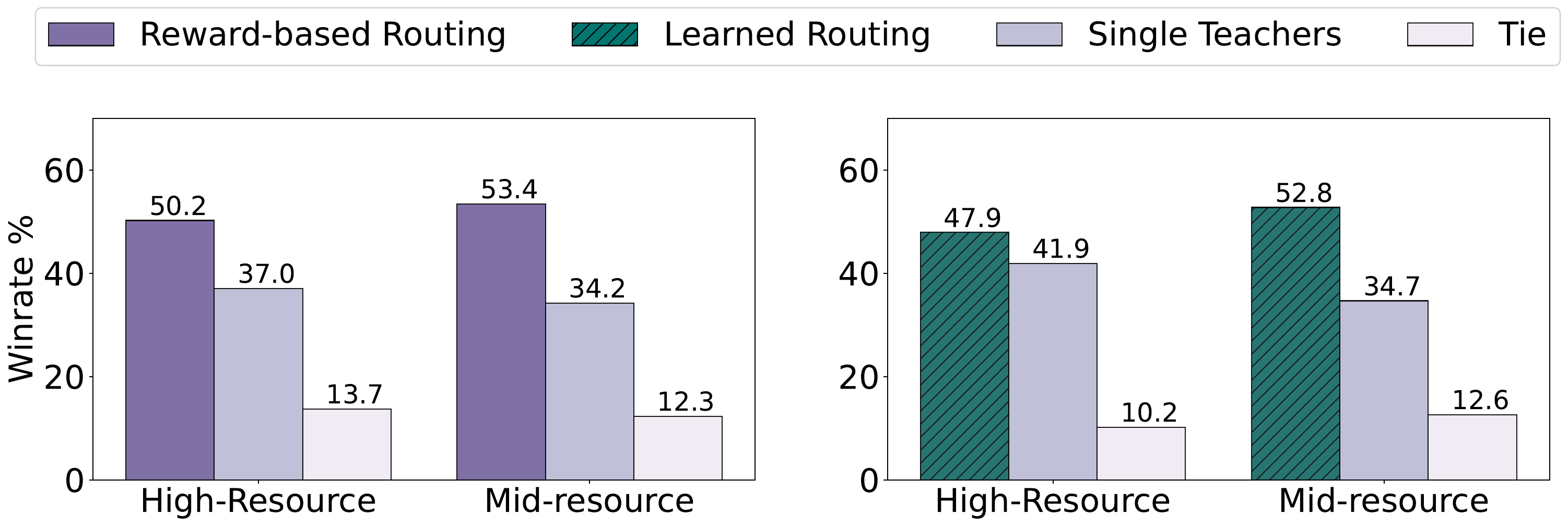}
        \caption{\textbf{Win rate Changes Across Language Resource Level:} Comparison of the Mid-Resource Languages and High-Resource Languages win rates against Single Teachers (results are the average of Aya 23, Llama 3 and Gemma 2 single teachers). Mid-resource languages consist of Turkish and Ukrainian and high-resource languages are English, German, French, Chinese and Arabic.}
        \label{fig:mid-res-vs-low-res}
\end{figure}

\textbf{Difference in per-language gains.}
Figure \ref{fig:mid-res-vs-low-res} shows performance gains for medium- versus high-resource languages when using reward-based and learned routing strategies compared to single teacher models such as Aya 23, Llama 3, and Gemma 2. 

Medium-resource languages, including Turkish and Ukrainian, experience greater benefits, with reward-based routing achieving a 56.1\% gain and learned routing achieving a 52.2\% gain over single teachers.
In contrast, high-resource languages~\citep{joshi2020state}, English, German, French, Chinese, and Arabic see a 35.7\% gain with reward-based routing and a 14.3\% gain with learned routing. These findings suggest that medium-resource languages gain more from routing strategies than from single teacher models. Detailed per-language gains are provided in Table~\ref{table:per_language_gains} in Appendix \ref{app:perLanguage}.

\begin{figure}[h!]
    \centering
         \includegraphics[width=0.9\textwidth]{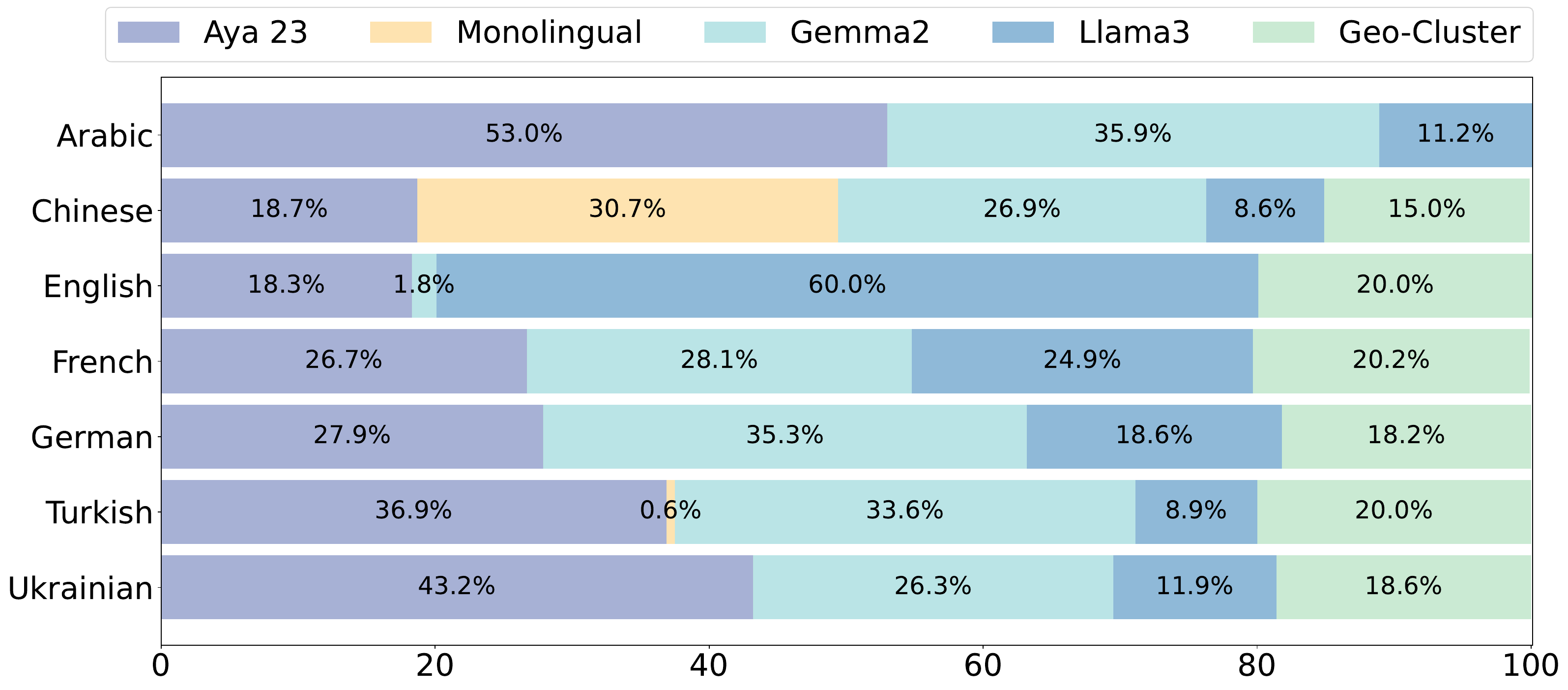}
         \caption{\textbf{Model Composition per Language:} Here we analyze the model routing distribution of a dataset constructed with Reward-Based Routing. The values represent the percentage of prompts routed to a given model for the particular language.}
         \label{fig:composition-rewardbased-routing}
\end{figure}

\textbf{Routed Dataset Distribution Across Models.} 
In Figure~\ref{fig:composition-rewardbased-routing}, we illustrate the distribution of the training dataset prompts routed to each model (for the reward-based router). We observed a balanced routing strategy with different models favored for each language, which highlights the benefits of combining the strengths of a pool of models with varying strengths. 
For instance, Llama 3, a strong English model, receives 60\% of English prompts but is less frequently used for other languages. Meanwhile, 30.7\% of Chinese prompts are directed to the Chinese monolingual expert, whereas the Turkish monolingual expert is rarely selected, with only 0.6\% of prompts routed to it.
Overall, Aya 23 emerges as the leading multilingual model, predominantly chosen for Ukrainian, Turkish, and Arabic, with 53\% of Arabic prompts routed to it. Geo-cluster models, included for all languages except Arabic (as there is no Geo-cluster model for it), handle an average of 18.7\% of the prompts.

\textbf{Comparison of in-language generation vs translation.}
In this section, we explore whether generating synthetic data directly in the target language is more effective than translating the best English-only data. To investigate this, we first generate English data using Llama 3 (the best English model), translate it into other 6 languages, and train a student model with this translated data.  We then compare this student model's performance to those trained with Llama 3's single-teacher generations and random-routing generations.

\begin{figure}[htb]
    \centering
         \includegraphics[width=0.8\textwidth]{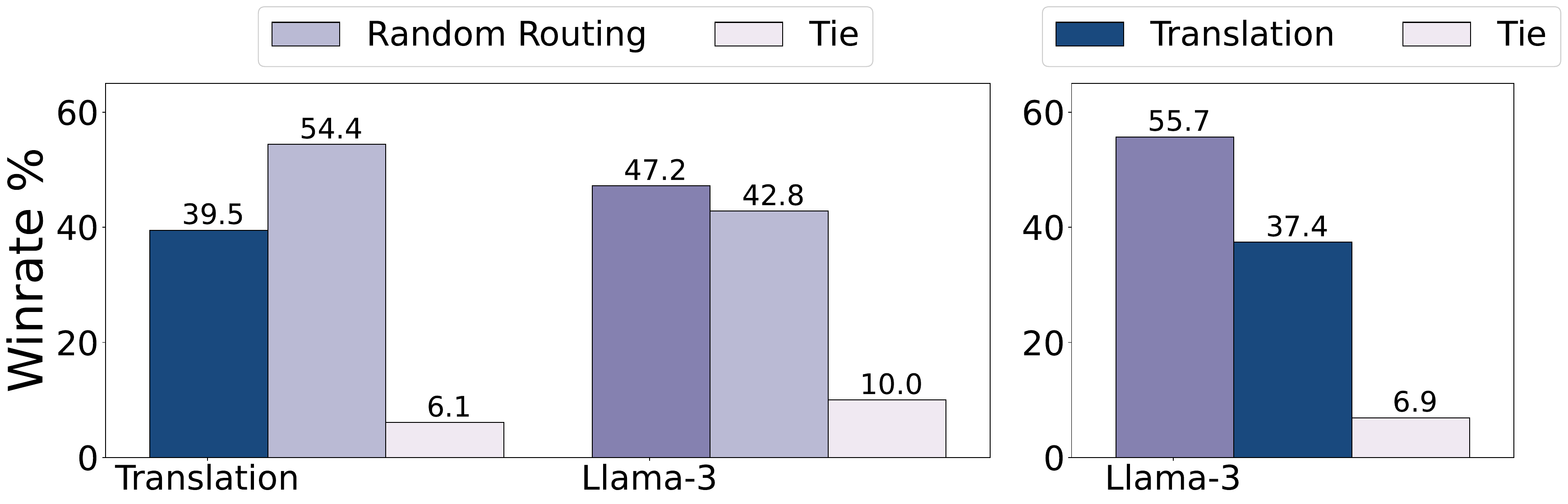}
         \caption{\textbf{Win rates \% of students trained with Llama 3 translations and Llama 3 generations:} Comparison of translation, in-language generation by single teacher and router-trained students to those trained with random routing. The largest gains are observed for in-language data generation with a win-loss diff of 18.3\%. All values are percentages and aggregated over 7 languages.}
         \label{fig:winrates-translation}
\end{figure}

Figure~\ref{fig:winrates-translation} demonstrates that random routing outperforms the translation baseline, achieving a win rate of 54.4\%, while the Llama 3 single teacher model exceeds the random-routing baseline with 10.3\% gain in this experiment.
Direct comparison of Llama 3 translation with Llama 3 single teacher students exhibits a significant 48.9\% increase in win rates for the single teacher model.
These results indicate that translation is the least effective method for synthetic data generation, as even random routing performs better. Generating samples within the original language offers substantial advantages over relying on single model translations, despite the single model being the top performer in the original language (English) before translation to others.

\subsection{Textual Characteristics}
\label{sec:textual_characteristics}

\begin{tcolorbox}[colframe=gray!90, colback=gray!20, title=Section Findings, fonttitle=\bfseries, boxrule=0.5pt, left=5pt, right=5pt, top=5pt, bottom=5pt, boxsep=5pt]
\begin{itemize}[left=0pt, topsep=2pt, itemsep=3pt, parsep=0pt]
    \item Routing strategies significantly increase the average number of tokens per generation, with outputs ranging from 160 to 242 tokens, compared to 76 tokens for the base model and around 144 tokens for both random routing and single teacher models, demonstrating that arbitrage methods produce longer text generations.
    \item The Gunning-Fog and Rix indices, which assess text readability and complexity, show higher scores for arbitrage methods compared to the all baselines. Specifically, The Gunning-Fog index increases by 3.28, 1.05 and 0.78 compared to the base student model, single teachers and random routing, respectively, indicating that arbitrage methods lead to more complex text compared to all baseline methods.
    \item The measure of lexical diversity (MLTD) is notably higher for arbitrage methods than for all baseline approaches. Reward-based routing achieves an increase of 7.97, and Learned Routing achieves an increase of 7.1 compared to the base student model. On average, arbitrage methods show absolute differences of 1.77 compared to single teacher models and 5.46 compared to random routing, highlighting their improved lexical diversity.
\end{itemize}
\end{tcolorbox}

To obtain a more holistic view of how multilingual arbitrage impacts model generation characteristics, we utilize the TextDescriptives framework from ~\citet{hansen2023textdescriptives} to calculate various textual features. We report average statistics, including the number of tokens along with readability and lexical diversity scores. Metrics like length are straightforward to compute and serve as positively correlated proxies for quality \citep{singh2024aya}. These metrics are calculated from model generations over 100 instances from the Dolly200 Eval set \citep{singh2024aya}. We standardize comparisons across models by allowing for a maximum output length of 600 tokens.

In addition to basic statistics like length, we also compute:
\begin{enumerate}
\item \textbf{Gunning Fog Index} \citep{Gunning1968TheTO} is a readability test that estimates the years of formal education required to understand a piece of text on the first reading. Gunning-Fog uses sentence length and prevalence of complex words to estimate the complexity of the text and assign a grade level between 0 and 20. A score of 17-18 indicates college graduate-level text.
\item \textbf{Rix} \citep{77e9b71c-fe84-3d01-afb5-fc072e945ca7} calculates readability based on the number of words with more than six characters divided by the number of sentences in the text. A score of 5 corresponds to a grade level of around 10, while a score of 7 or higher indicates the need for a higher educational level to comprehend.
\item \textbf{Measure of Textual Lexical Diversity (MTLD) score}~\citep{lex} helps tracking changes in vocabulary by reflecting the average number of words in a sequence that maintains a certain type-token ratio (TTR), a measure of vocabulary variety~\citep{McCarthy2010MTLDVA}. An MLTD score of 50 can be considered as moderate lexical diversity.
\end{enumerate}

We present all the results in Figure~\ref{fig:text_char} and Table~\ref{tab:textualcharacteristics-results} in Appendix~\ref{app:textChar}. 
\begin{figure}[htb]
    \centering
         \includegraphics[width=0.8\textwidth]{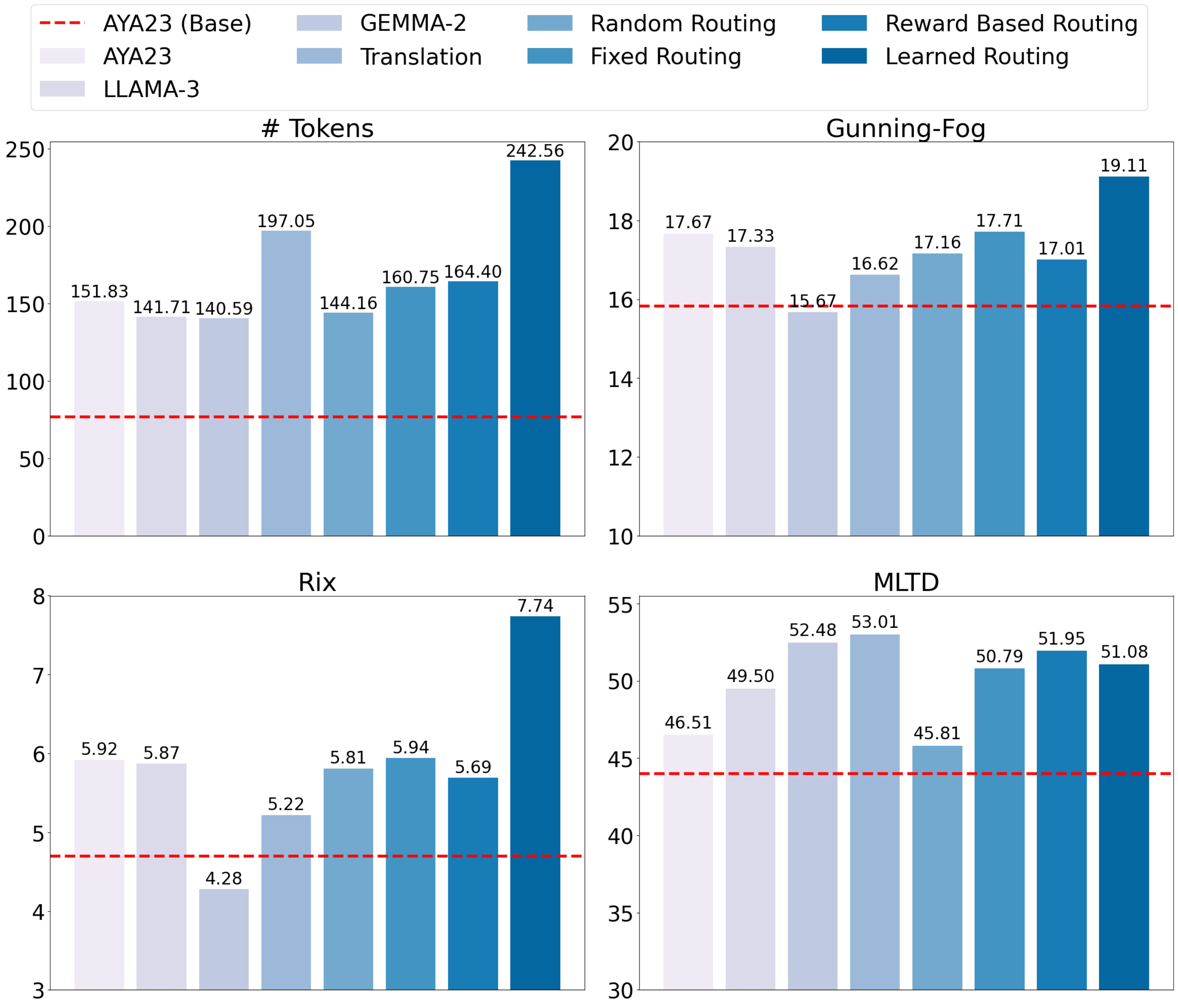}
         \caption{\textbf{Evaluation of Textual Characteristics:} We analyze characteristics of student models in four languages: \textsc{English, German, French and Ukranian}. The number of tokens, Gunning-Fog, Rix Index, and MLTD for each model highlights the differences in verbosity, readability and lexical diversity.}
         \label{fig:text_char}
\end{figure}

\textbf{Average number of tokens per generation.} 
The most significant change is observed in the average number of tokens per generation. The base model generates an average of 76 tokens per generation, whereas routing approaches produce substantially longer outputs, ranging from 160 tokens with Fixed Routing to 242 tokens with Learned Routing. In contrast, both random routing and single teacher models (averaged across Aya 23, Llama 3, and Gemma 2) generate around 144 tokens on average. These findings demonstrate that arbitrage methods result in longer text generations compared to both random routing and single teacher models.

\textbf{Textual properties.} 
The readability metrics show smaller absolute changes compared to the average number of tokens. For the Gunning-Fog index, changes range from a decrease of 0.16 for Gemma 2 to an increase of 3.28 for Learned Routing, relative to the base student model. Similarly, the Rix index varies from a decrease of 0.42 for Gemma 2 to an increase of 3.04 for Learned Routing. Both metrics reveal that arbitrage methods result in higher scores. The Gunning-Fog index shows an absolute difference of 1.05 between arbitrage methods and single teacher models, whereas the difference is 0.78 for random routing. For the Rix index, the absolute difference is 1.11 between arbitrage methods and single teachers, compared to 0.65 for random routing.

These indices serve as proxies for evaluating text complexity. There is a clear trend indicating that multilingual arbitrage strategies, especially the learned routing approach, lead to higher readability metrics. In contrast, single teacher models, especially Gemma 2, generally result in lower values.

Regarding the MLTD score, we observe significant changes, with Reward-based routing showing an increase of up to 7.97 and Learned routing showing an increase of 7.1 relative to the base student model, which are considered substantial improvements~\citep{10.1093/applin/amw009}.
Arbitrage methods result in higher MLTD scores compared to both random routing and single teacher results. The average absolute difference is 1.77 between arbitrage methods (averaged over all 3 methods) and single teacher models (averaged over Aya 23, Llama 3 and Gemma 2), while the difference is 5.46 for random routing.

Overall, multilingual arbitrage strategies significantly increase the number of tokens in generations, readability metrics and improve lexical diversity compared to single teacher models. This suggests that multilingual arbitrage enhances data quality and diversity, which in turn leads to improvements in student model performance and explains the significant increase in win rates.

\textbf{Routed Dataset Composition Characteristics.}

\begin{figure}[ht!]
    \centering
    \begin{subfigure}[b]{0.48\textwidth}
        \centering
        \includegraphics[width=\textwidth]{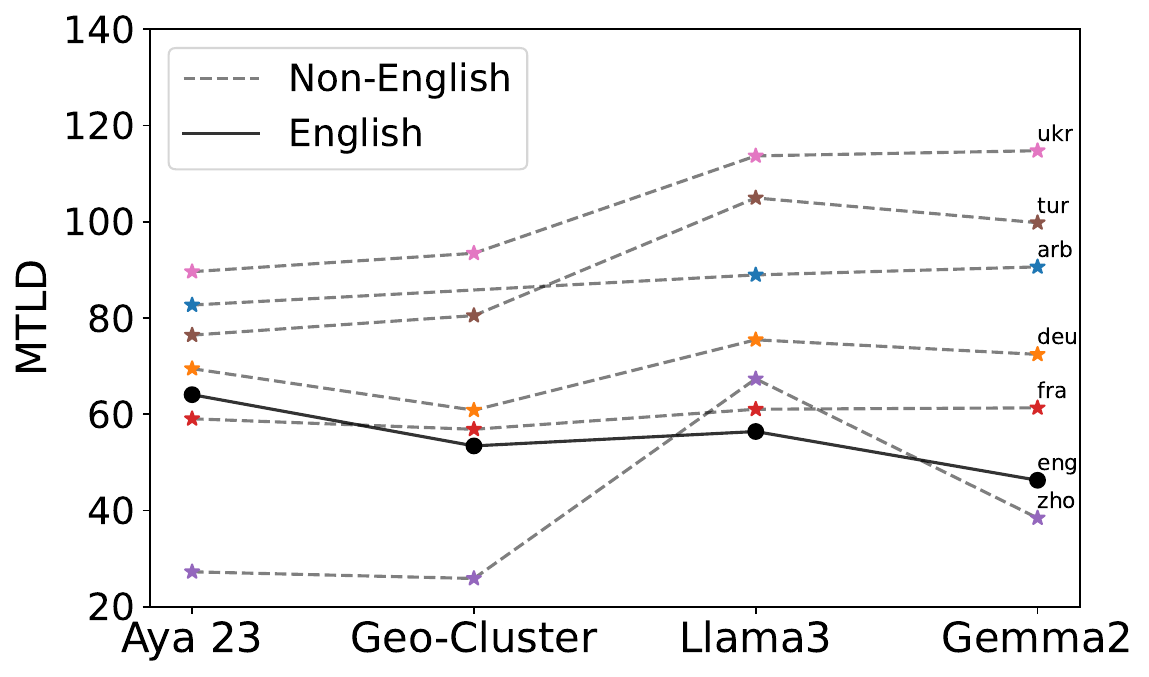}
        \caption{Average MTLD per model.}
        \label{fig:mtld-rewardbased-routing}
    \end{subfigure}
    \hfill
    \begin{subfigure}[b]{0.48\textwidth}
        \centering
        \includegraphics[width=\textwidth]{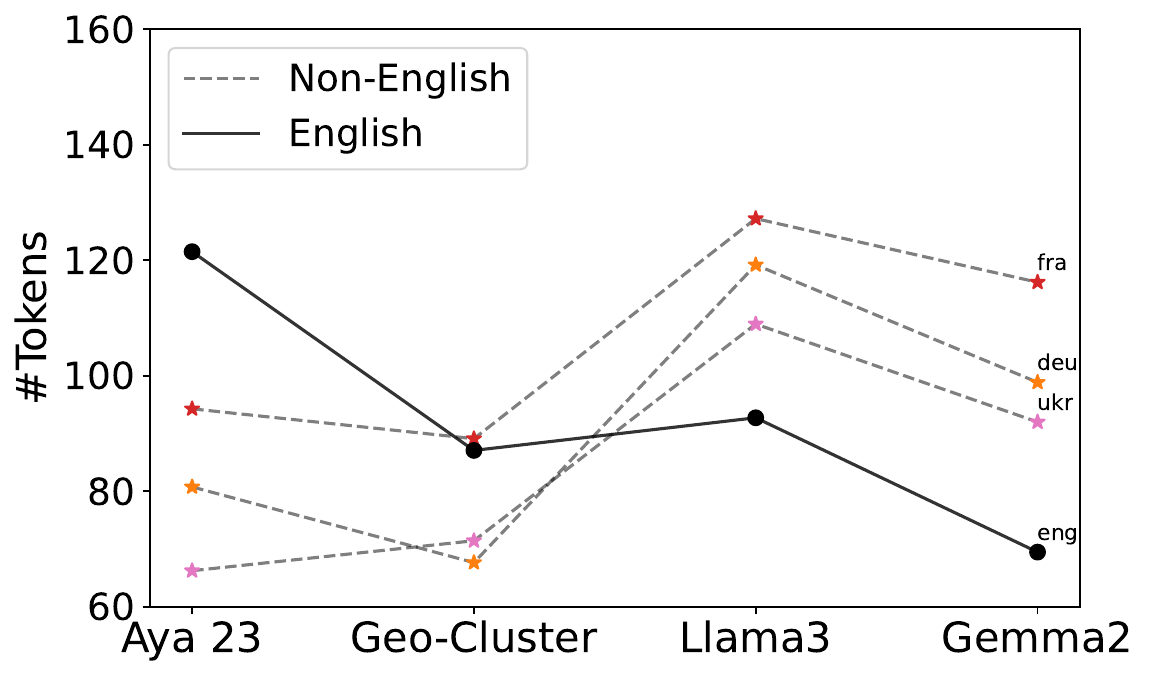}
        \caption{Average number of tokens per model.}
        \label{fig:ntokens-rewardbased-routing}
    \end{subfigure}
    \caption{\textbf{Characteristics of Prompts Routed to Given Models:} We analyze the MTLD (a) and number of tokens (b) for the set of prompts routed to each of the teacher models as selected by Reward-Based Routing. Each line represents a different language and each column is a particular teacher model.}
    \label{fig:routing-characteristics}
\end{figure}

Here, we analyze how prompt characteristics affect the reward-based router decision, using the same subset of the UltraFeedback Binarized Dataset as depicted in Figure~\ref{fig:composition-rewardbased-routing}. The average MLTD score and number of tokens of the prompts routed to a particular model is shown in Figure~\ref{fig:routing-characteristics}. 

Figure~\ref{fig:mtld-rewardbased-routing} shows that the average MLTD scores \emph{for English prompts} routed to different models range from 46.28 to 64.07. Aya 23 receives English prompts with the highest MLTD score of 64.07, while Llama 3 has an average MLTD score of 56.41, and Gemma 2 has the lowest score of 46.28. In contrast, for non-English prompts, Aya 23 has an average MLTD score of 67.42, Llama 3 scores 79.66, and Gemma 2 achieves the highest MLTD score of 85.24.

Figure~\ref{fig:ntokens-rewardbased-routing} shows that the longest English prompts are routed to Aya 23, with an average of 121.5 tokens, while Gemma 2 receives the shortest English prompts, averaging 69.4 tokens. English prompts routed to Geo-clusters and Llama 3 have average token counts of 87.1 and 92.7, respectively.
\emph{For non-English prompts, the pattern differs.} Geo-clusters receive the shortest prompts, averaging 78.8 tokens. Aya 23 receives prompts with an average of 90.7 tokens, Gemma 2 with 94.1 tokens, and Llama 3 receives the longest non-English prompts, averaging 112.0 tokens.

We can conclude, for English prompts, those that are more lexically diverse and longer tend to be routed to Aya 23. 
In contrast, for non-English prompts, Gemma 2 and Llama 3 are preferred for handling more lexically diverse and longer prompts.

\section{Related Work \label{sec:related_work}}

\textbf{LLM circularity.}
The issue of LLM circularity, where models influence other LLMs through distilled data, has recently gained attention. Research to date has focused on two main areas: model degradation via recursive training \citep{dohmatob2024model, briesch2023large, shumailov2023curse} and self-preference in an LLM-as-a-Judge setting. Regarding model degradation, studies have shown that training LLMs with data iteratively generated by other LLMs impairs performance as the tails of the original distribution start to disappear. This includes work focusing solely on high-frequency contexts, thereby neglecting long-tail knowledge \citep{briesch2023large, bertrand2024stability, shumailov2024curse, dohmatob2024tale}, and resulting in a loss of diversity \citep{guo2024curious, feng2024beyond}.
In \citep{shimabucoro2024llmseellmdo}, the authors explore how the transfer of characteristics via passive inheritance occurs when synthetic data generated by different LLMs is involved. By considering the issues highlighted in these studies, we aim to optimize synthetic data generation by selecting the most calibrated teacher model from a pool of LLMs in a multilingual setting. 

\textbf{Instruction Fine-tuning (IFT).}
IFT on a large collection of tasks has become a key paradigm for enhancing LLM performance, increasing their utility \citep{sanh2021multitask,wei2021finetuned,mishra2021cross,min2021metaicl,ouyang2022training}, and enabling generalization to unseen tasks \citep{wei2022finetunedlanguagemodelszeroshot,chung2022scaling}.
Successful instruction tuning relies on three critical factors: task diversity \citep{longpre2023flan,wang2023far,chung2022scaling}, complexity \citep{xu2023wizardlm,luo2023wizardcoder,luo2023wizardmath}, and quality \citep{zhou2023lima,taori2023stanford,muennighoff2023octopack,zhuo2024astraios}.
While these principles have been validated primarily for English tasks, there is a growing focus on instruction fine-tuning in multilingual contexts \citep{ustun2024aya, muennighoff2022crosslingual}. A common challenge is the scarcity of multilingual instruction datasets, which recent efforts \citep{singh2024aya} have addressed by creating a comprehensive multilingual instruction dataset. In this work, we strategically sample from a diverse pool of models, each with unique strengths across different languages, to generate high-quality synthetic instruction data.

\textbf{Multilingual Synthetic Data.} 
While numerous studies focus on synthetic data generation in English settings \citep{gao2023selfguidednoisefreedatageneration, anabytavor2019datadeeplearningrescue, wang2022self, alpaca, gao2023selfguidednoisefreedatageneration}, the impact of synthetic data on enhancing multilingual performance is not well understood \citep{kaddour2023text, yadav2023assessing, kramchaninova2022synthetic}.
Recently, \citep{aryabumi2024aya} explored multilingual synthetic data generation in the context of a single teacher model.
\citet{aakanksha2024multilingualalignmentprismaligning} used multilingual synthetic data for safety preference training, and \citep{dang2024rlhfspeaklanguagesunlocking} applied it for multilingual preference tuning. Our research diverges by concentrating on multilingual synthetic instruction data generation from an ecosystem view rather than a single teacher. 

\textbf{Large Language Model Ensemble.} 
Ensembling LLMs is a powerful approach to leverage individual model strengths while mitigating weaknesses through complementary abilities. The aim is to combine available LLMs to enhance performance consistently across various downstream tasks. There is limited research on the complementary potential of LLMs and strategies for assembling them effectively. \citet{jiang2023llm} introduce a framework that combines LLMs using pairwise ranking and generative fusion to leverage their diverse strengths. \citet{chen2023frugalgpt} use a sequential inference approach with existing LLMs, continuing until the response quality is satisfactory. \citet{wang2023fusing} tackle the challenge of fusing outputs from expert models with complementary data distribution knowledge, framing it as a supervised learning problem. \citet{shnitzer2023large} develop a router to select the best LLM candidate based on benchmark evaluations. These works focus on LLM ensembles in a primarily monolingual, mostly English setting. The most relevant work to ours is~\citep{lu-etal-2024-routing}, which proposes a reward model-guided routing method to utilize complementary model strengths across tasks. In contrast, our work explores various routing strategies beyond reward-based routing, focusing on multilingual LLMs in multilingual contexts.

\section{Conclusion\label{sec:conclusion}}
\label{sec:conclusion}

In this work, we introduce the concept of  \textit{multilingual arbitrage}, which strategically utilizes performance variations across different models for a given language to sample from a pool of teacher models, thereby generating a superior dataset for training effective student models. Our extensive experiments across 15 languages demonstrate the efficacy of our routing strategies, showing that this approach significantly enhances student models' performance over all benchmarks of interest. We observe notable gains in open-ended generation tasks and discriminative benchmarks compared to the traditional single-teacher data generation and training method. Furthermore, through additional analysis of textual characteristics and evaluation on unseen discriminative tasks, we confirm that our instruction fine-tuned students not only retain their initial capabilities but also improve their multilingual generation skills. Our work motivates the merit of strategic sampling, particularly where a diverse pool of models is expected to excel at different parts of the data distribution. We expect \textit{arbitrage} techniques to provide sizable gains when addressing out-of-distribution problems, rare or often underrepresented parts of the data distribution.

\section{Limitations\label{sec:limitations}}

One of the limitations of our work is we do not evaluate the impact of \textit{multilingual arbitrage} on safety. Furthermore, while we consider a wide pool of teachers comprised of very distinct model characteristic -- our pool of teachers are all in a similar scale of 8 billion parameters. The focus of models in the same size classes was forced by the large computational cost of our experimental set-up. We leave as an important topic of future work exploring the impactof scaling on the gains we observe.

\bibliography{paper}

\appendix
\section{Appendix}
\subsection{Teacher Model Pool Details}
\label{app:teacher_pool}

\textbf{Single Teacher Models.}
We include additional details about each of the single teacher models we benchmark below:

\begin{itemize}
\item \textbf{Aya-23-8B}~\citep{aryabumi2024aya} is an 8B parameter model and a part of the Aya-23 family of multilingual instruction-tuned language models that supports 23 languages, and are based on Cohere's Command model\footnote{\url{https://cohere.com/command}} and multilingual instruction-style collection~\citep{singh2024aya}.

\item \textbf{Llama-3-8B-instruct}~\citep{dubey2024llama} is an open-source instruction-tuned version of the Llama-3-8B pre-trained model. The model is trained on over 15 trillion tokens of publicly available data, with a focus on optimizing the performance across various real-world scenarios, including reasoning and code generation.

\item \textbf{Gemma-2-9B-it}~\citep{team2024gemma} is a 9B parameter instruction fine-tuned model on 8T tokens of data from web documents, code, and science articles. In particular, the 9B model was trained with knowledge distillation~\citep{hinton2015distillingknowledgeneuralnetwork} instead of next token prediction.
\end{itemize}

\textbf{Monolingual Teacher Models.}
These models are specifically tailored for individual languages, specifically Chinese and Turkish:
\begin{itemize}
\item \textbf{Qwen2-7B-instruct}~\citep{yang2024qwen2technicalreport} is an open-source 7B parameter model pretrained on 7T tokens of data from code, mathematics, and multilingual data. Qwen2-7B-instruct is a multilingual model supporting approximately 30 languages, and showing strong performance on Chinese.
\item \textbf{Turkish-Llama-8b-Instruct-v0.1}\footnote{\url{https://huggingface.co/ytu-ce-cosmos/Turkish-Llama-8b-Instruct-v0.1}} is a fully fine-tuned version of the Llama-3-8B-instruct model with a 30GB Turkish dataset. It currently tops the Turkish leaderboard on HuggingFace\footnote{\url{https://huggingface.co/spaces/malhajar/OpenLLMTurkishLeaderboard_v0.2}}  for text generation tasks.
\end{itemize}

\subsubsection{Geo-Cluster Training Details}
\label{app:geocluster_training}

\begin{table}[h!]
    \centering
    \scalebox{0.8}{
    \begin{tabular}{lcccc} 
        \toprule
        & \multicolumn{4}{c}{\textbf{Number of Samples Per Dataset}} \\
        \cmidrule{2-5}
        \textbf{Language Cluster} & Original ShareGPT & ShareGPT CommandR+ & Original Dolly15k & Dolly15k CommandR+ \\
        \midrule
        \textsc{Germanic} & 155,480 & 157,699 & 40,466 & 42,447 \\
        \noalign{\smallskip} 
        \textsc{Slavic} & 259,217 & 263,488 & 67,721 & 71,121 \\
        \noalign{\smallskip} 
        \textsc{Romance} & 309,708 & 314,513 & 80,295 & 84,345 \\
        \noalign{\smallskip} 
        \textsc{East Asian} & 230,848 & 235,369 & 58,864 & 61,743 \\
        \bottomrule
    \end{tabular}
    }
    \caption{Number of Training Samples Per Language Cluster}
    \label{tab:data-distribution}
\end{table}

To train highly performant Geo-clusters, we train an 8B parameter Cohere command model \footnote{\url{https://cohere.com/command}} on a data mix of the 15 languages covered by the Geo-Clusters as shown in Table~\ref{table:clusters}. For this data mix, we used both ShareGPT dataset and the Dolly-15k dataset as described by \citep{aryabumi2024aya}. First these two datasets' prompts and completions were translated into these 15 languages, and translations were done using the NLLB-3.3B model~\citep{costa2022no}. In addition, we also included what we call the ShareGPT CommandR+ dataset and the Dolly-15k CommandR+ dataset. For these variants, we use the translated prompts generated completions for the translated prompts using Command R+\footnote{\url{https://huggingface.co/CohereForAI/c4ai-command-r-plus}}. Our datasets cover 15 languages shown in Table~\ref{table:clusters}. Table~\ref{tab:data-distribution} shows the training data distribution in terms of number of samples used for each Geo-Cluster model training.

\begin{figure}[htb]
    \centering
         \includegraphics[width=0.9\textwidth]
         {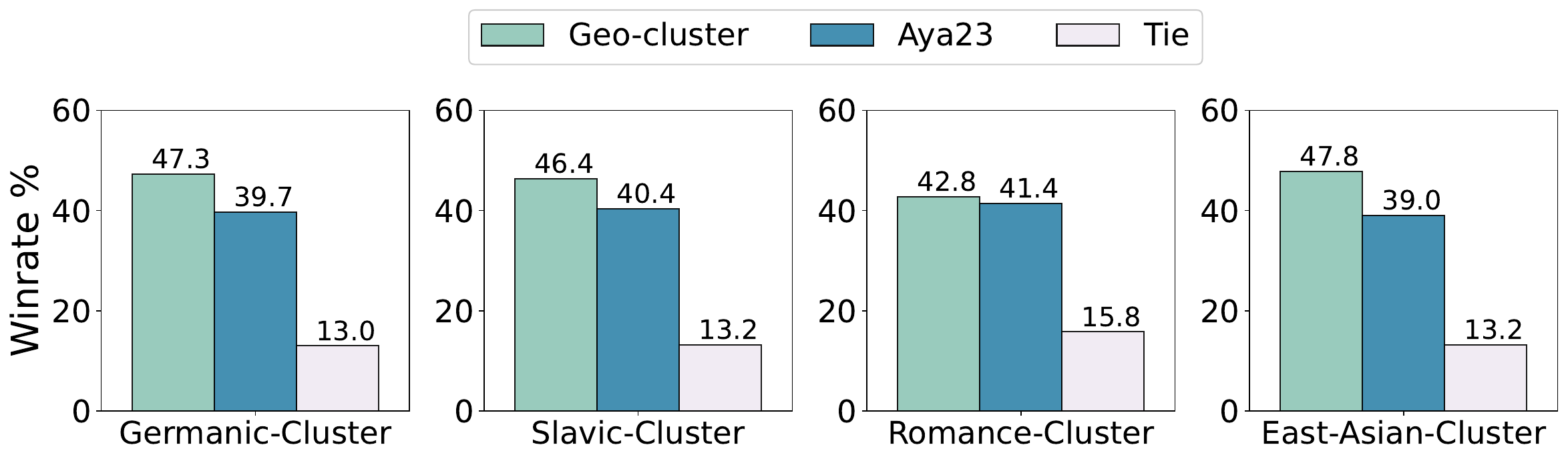}
         \caption{Geo-cluster win-rates against Aya 23 Single Teacher Model after training. All values are percentages, and aggregated over number of languages in each language cluster. Geo-cluster are powerful teacher models relative to the capabilities of the base Aya model.} 
        \label{fig:geoclusters-winrates}
\end{figure}

Before using the geo-clusters as teacher models, we validate performance of our trained Geo-cluster models. We compute average win rates in each language cluster using the held-out multilingual Dolly-200 evaluation dataset~\citep{ustun2024aya}. 

\begin{table*}[th]
\centering
\small
\begin{tabular}{ll}
\toprule
\textbf{Language} & \textbf{Model Pool} \\
\midrule
\textsc{Arabic} & Base Pool \\
\noalign{\smallskip} 
\textsc{Chinese} & Base Pool, East Asian + Turkish Cluster, Qwen2-7B \\
\noalign{\smallskip} 
\textsc{English} & Base Pool, Germanic Cluster \\
\noalign{\smallskip} 
\textsc{French} & Base Pool, Romance Cluster \\
\noalign{\smallskip} 
\textsc{German} & Base Pool, Germanic Cluster \\
\noalign{\smallskip} 
\textsc{Turkish} & Base Pool, East Asian + Turkish Cluster, Turkish-Llama-8b \\
\noalign{\smallskip} 
\textsc{Ukrainian} & Base Pool, Slavic Cluster \\
\bottomrule
\end{tabular}
\caption{Teacher model pool available for each language. The \emph{Base Pool} consists of those outlined in Section \ref{subsec:single_teacher}: Aya 23, Llama 3, Gemma 2.}
\label{tab:model_pool}
\end{table*}

\subsection{Language Families}
\label{app:language_fam}

As we present in Section~\ref{sec:student_models}, we generate synthetic data in seven diverse languages: \emph{Arabic, Chinese, English, French, German, Turkish}, and \emph{Ukrainian}. These languages, representing different language families, are selected to ensure a comprehensive evaluation across various linguistic contexts, detailed in Table~\ref{table:language_lineage}. 
\begin{table}[h!]
\centering
\begin{tabular}{lllllc}
\toprule
\textbf{ISO} \textbf{Code} & \textbf{Language} & \textbf{Script} & \textbf{Family} & \textbf{Subgrouping} & \textbf{Resources} \\
\midrule
ara & Arabic      & Arabic       & Afro-Asiatic   & Semitic           & High \\
zho & Chinese     & Han          & Sino-Tibetan   & Sinitic           & High \\
eng & English     & Latin        & Indo-European  & Germanic          & High \\
fra & French      & Latin        & Indo-European  & Italic            & High \\
deu & German      & Latin        & Indo-European  & Germanic          & High \\
tur & Turkish     & Latin        & Turkic         & Common Turkic     & Mid \\
ukr & Ukrainian   & Cyrillic     & Indo-European  & Balto-Slavic      & Mid \\
\bottomrule
\end{tabular}
\caption{\textbf{Lineage for Cluster Languages.} 7 languages covered by our main experiments, each language’s corresponding script, family, subgrouping, and if it is classified as higher or mid-resourced according to~\citep{joshi2020state}.}
\label{table:language_lineage}
\end{table}

\subsection{Router Model Details \label{app:router-model-details}}

\textbf{Training Details.}
We chose Gemma2-2B\footnote{\url{https://huggingface.co/google/gemma-2-2b}} as our router model for its compact size, performance, and multilingual capabilities. We fine-tuned Gemma2-2B model using the AdamW~\citep{loshchilov2019decoupledweightdecayregularization} optimizer with an initial learning rate of $5 \times 10^{-5}$. We used a linear learning rate scheduler with a 200 warmup steps. We set weight decay to $0$ and fine-tuned for 2 epochs. 

To further improve training efficiency, we also evaluate a smaller mT5-base\footnote{\url{https://huggingface.co/google/mt5-base}} variant with 580M parameters. We finetuned the mT5-base using the Adafactor\footnote{\url{https://discuss.huggingface.co/t/how-is-the-adafactorscheluder-suppose-to-be-used/9007}} optimizer with $1 \times 10^{-3}$ as the learning rate. We fine-tuned for 5 epochs with a train batch size of 32.

\textbf{Comparison of mT5 and Gemma 2 as Router Model.}
We chose Gemma2-2B as the final candidate for our learned router model. The student model trained on the dataset routed by Gemma2-2B demonstrated significant improvements, particularly against the strong Gemma2-9B single teacher model. Gemma2-2B was used as the learned router in all our experiments. 

\begin{figure}[h!]
    \centering
    \includegraphics[width=1.0\textwidth]{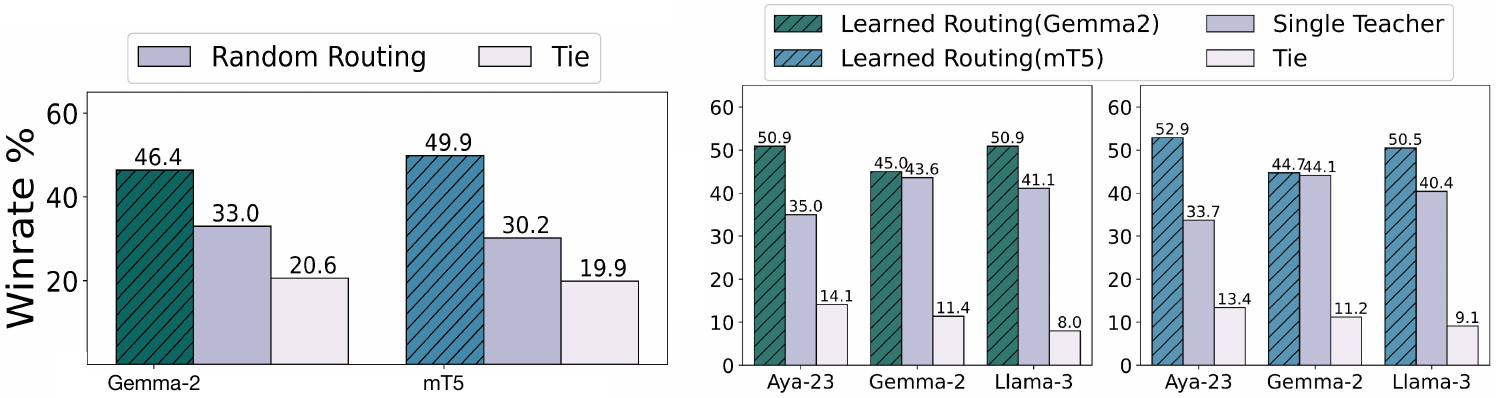}
    \caption{Win-rate \% comparison of Learned Routing (mT5) and Learned Routing (Gemma2) against Random Routing  (left) and multiple Single Teacher Models (right).}
    \label{fig:router-comparison}
\end{figure}

Figure~\ref{fig:router-comparison} shows Gemma2-2B and mT5-base router performances compared to random routing and single teachers. Despite its smaller size, mT5-base also achieved remarkable results, outperforming all baseline approaches with a notable 65.2\% gain over random routing and an average gain of 27.7\% over single teacher models.

\subsection{Difference in per-language gains.}
\label{app:perLanguage}

\begin{figure}[h!]
    \centering
         \includegraphics[width=0.9\textwidth]{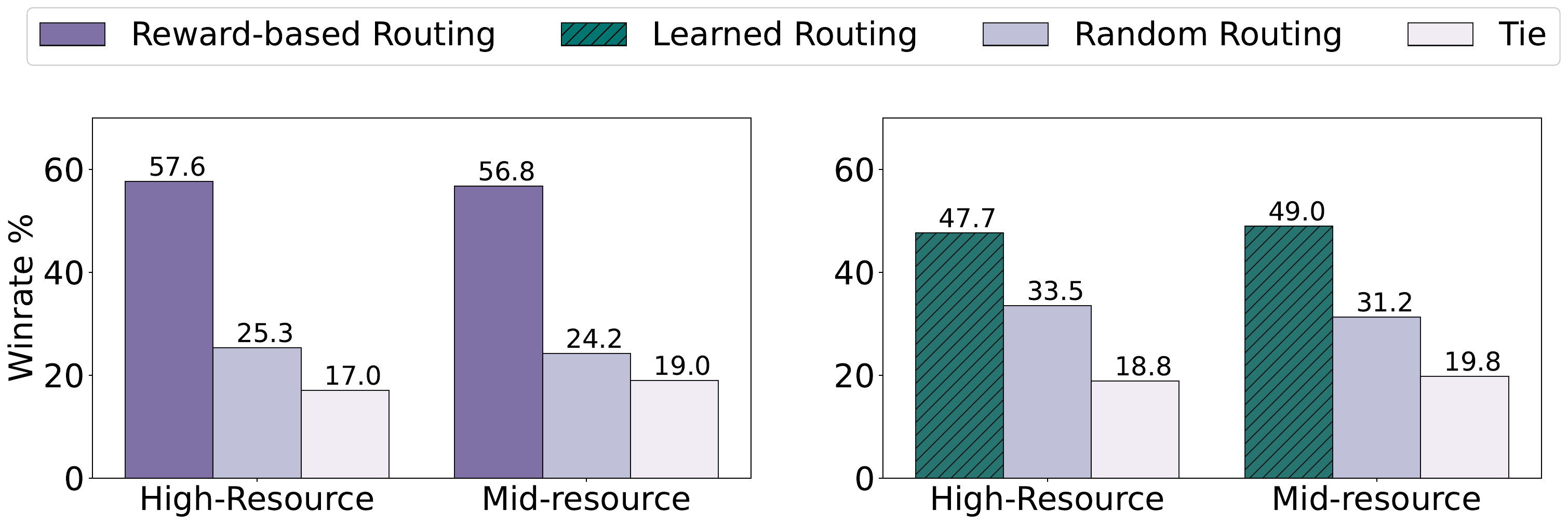}
        \caption{\textbf{Win-rate Changes Across Language Resource Level.} We compare the win rates of Mid-Resource Languages and High-Resource Languages against random-routing. Mid-resource languages consist of Turkish and Ukrainian and high-resource languages are English, German, French, Chinese and Arabic.}
        \label{fig:mid-res-vs-low-res-rnd}
\end{figure}

In Figure \ref{fig:mid-res-vs-low-res-rnd}, we compare both reward-based routing and learned routing strategies against random routing for medium-resource and high-resource languages.

High-resource languages~\citep{joshi2020state}, English, German, French, Chinese, and Arabic see a 127.6\% gain with reward-based routing and a 42.4\% gain with learned routing. Medium-resource languages that includes Turkish and Ukrainian, experience greater benefits, with reward-based routing achieving a 134.7\% gain and learned routing achieving a 57.1\% gain over random routing.
These findings suggest that medium-resource languages gain more from strategic sampling than from random routing. Detailed per-language gains are provided in Table~\ref{table:per_language_gains}.

\begin{table}[h!]
\centering
\begin{tabular}{lccccc}
\toprule
& \multicolumn{2}{c}{\textbf{\% gain (Single Teachers)}} & \multicolumn{2}{c}{\textbf{\% gain (Random Routing)}} \\
\cmidrule{2-5}
\textbf{Language} & \textbf{Reward-based} & \textbf{Learned} & \textbf{Reward-based} & \textbf{Learned} \\
\midrule
Arabic      & \textcolor{blue}{75.7}  &  43.4  &  115.1  & 43.5      \\
Chinese     & \textbf{114.5} &  2.9   &  101.8  & -4.6      \\
English     & 55.2     &  0.4   &  116.0  & \textbf{115.7}  \\
French      & 22.5           &  -4.4  &  79.3   &  39.1     \\
German      & 31.7           &  28.8  &  76.7  &  88.7     \\
Turkish     & 52.2 & \textbf{59.6} & \textbf{228.9} & \textcolor{blue}{94.5}  \\
Ukrainian   & 59.9  &  \textcolor{blue}{43.7}  &  \textcolor{blue}{172.9}  &  87.2     \\
\bottomrule
\end{tabular}
\caption{\textbf{Win-rate gains across languages.} This table presents the percentage gain of reward-based routing and learned routing compared to single teachers and random routing across seven languages. The highest gain in each column is highlighted in \textbf{bold}, while the second highest gain is indicated in \textcolor{blue}{blue}.}
\label{table:per_language_gains}
\end{table}

The results indicate that reward-based routing leads to larger gains across all languages compared to learned routing, whether against single teachers or random routing. Mid-resource languages, Turkish and Ukrainian, consistently show high gains in all scenarios, followed by Arabic. However, no distinct pattern emerges for high-resource languages. Notably, reward-based routing results in significant gains for Chinese against both random routing and single teachers. Additionally, both reward-based and learned routing achieve substantial gains for English when compared to random routing.

\pagebreak

\subsection{Discriminative tasks.}
\label{app:discTasks}

\begin{table}[ht!]
    \centering
    \scalebox{0.9}{
    \begin{tabular}{lcccc}
        \toprule
         & \textbf{XCOPA} & \textbf{XNLI} & \textbf{XStoryCloze} & \textbf{Average} \\
        \midrule
        \textsc{\textbf{Base Model}} \\
        \noalign{\smallskip} 
        \textsc{AYA23} (Base) & 64.1 &  42.9 &  68.23 &   58.41 \\ [3mm]
        \textsc{\textbf{Single Teacher}} \\
        \noalign{\smallskip} 
        \textsc{AYA23}  & 65.5 \tcbox[colback=LightGreen]{\textcolor{black}{$\uparrow 2.18$}} & 43.86 \tcbox[colback=LightGreen]{\textcolor{black}{$\uparrow 2.23$}} & 68.05 \tcbox[colback=LightRed]{\textcolor{black}{$\downarrow 0.27$}} &  59.13 \tcbox[colback=LightGreen]{\textcolor{black}{$\uparrow 1.23$}} \\
        \textsc{LLAMA-3} & 65.1 \tcbox[colback=LightGreen]{\textcolor{black}{$\uparrow 1.56$}} & 44.04 \tcbox[colback=LightGreen]{\textcolor{black}{$\uparrow 2.65$}} & 66.46 \tcbox[colback=LightRed]{\textcolor{black}{$\downarrow 2.60$}} & 58.53 \tcbox[colback=LightGreen]{\textcolor{black}{$\uparrow 0.20$}} \\
        \noalign{\smallskip}
        \textsc{GEMMA-2} & 66.1 \tcbox[colback=LightGreen]{\textcolor{black}{$\uparrow 3.12$}} & 43.98 \tcbox[colback=LightGreen]{\textcolor{black}{$\uparrow 2.51$}} & 67.74 \tcbox[colback=LightRed]{\textcolor{black}{$\downarrow 0.72$}} & 59.3 \tcbox[colback=LightGreen]{\textcolor{black}{$\uparrow 1.52$}} \\
        \noalign{\smallskip} 
        \textsc{Translation} & 64.6 \tcbox[colback=LightGreen]{\textcolor{black}{$\uparrow 0.78$}} & 43.46 \tcbox[colback=LightGreen]{\textcolor{black}{$\uparrow 1.30$}} & 66.77 \tcbox[colback=LightRed]{\textcolor{black}{$\downarrow 2.14$}} & 58.27 \tcbox[colback=LightRed]{\textcolor{black}{$\downarrow 0.24$}} \\ [3mm]
        \noalign{\smallskip} 
        \textsc{\textbf{Multilingual Arbitrage}} \\
        \noalign{\smallskip}         
        \textsc{Random Routing} & 65.9 \tcbox[colback=LightGreen]{\textcolor{black}{$\uparrow 2.80$}} & 44.01 \tcbox[colback=LightGreen]{\textcolor{black}{$\uparrow 2.58$}} & 67.25 \tcbox[colback=LightRed]{\textcolor{black}{$\downarrow 1.44$}} &  59.05 \tcbox[colback=LightGreen]{\textcolor{black}{$\uparrow 1.09$}}\\
        \noalign{\smallskip}
        \textsc{Fixed Routing} & \textbf{67.4}
        \tcbox[colback=LightGreen]{\textcolor{black}{$\uparrow 5.14$}} & 43.89
        \tcbox[colback=LightGreen]
        {\textcolor{black}{$\uparrow 2.30$}} & 68.33
        \tcbox[colback=LightGreen]
        {\textcolor{black}{$\uparrow 0.14$}} & \textbf{59.87}
        \tcbox[colback=LightGreen]{\textcolor{black}{$\uparrow 2.50$}}\\
        \noalign{\smallskip}
        \textsc{Reward Based Routing} & 66.2
        \tcbox[colback=LightGreen]{\textcolor{black}{$\uparrow 3.27$}} & \textbf{44.21} \tcbox[colback=LightGreen]{\textcolor{black}{$\uparrow 3.05$}} & 68.20 \tcbox[colback=LightRed]{\textcolor{black}{$\downarrow 0.05$}} &  59.53 \tcbox[colback=LightGreen]{\textcolor{black}{$\uparrow 1.91$}} \\
        \noalign{\smallskip} 
        \textsc{Learned Router} & 65.8 \tcbox[colback=LightGreen]{\textcolor{black}{$\uparrow 2.65$}} & 43.62 \tcbox[colback=LightGreen]{\textcolor{black}{$\uparrow 1.67$}} & \textbf{68.36} \tcbox[colback=LightGreen]{\textcolor{black}{$\uparrow 0.19$}} &  59.25
        \tcbox[colback=LightGreen]{\textcolor{black}{$\uparrow 1.43$}} \\
        \noalign{\smallskip}
        \bottomrule
    \end{tabular}}
    \caption{Performance of Student Models on held-out Discriminative Tasks: XCOPA, XNLI, and XStoryCloze. The results are averaged over seven languages, highlighting the improvements or declines in performance compared to the base model AYA23.}
    \label{tab:discriminative-results}
\end{table}

\subsection{Textual Characteristics}
\label{app:textChar}

\begin{table}[ht!]
    \centering
    \scalebox{0.9}{
    \begin{tabular}{lcccc}
        \toprule
        \textbf{Student Models}  & \textbf{\# Tokens}  &  \textbf{Gunning-Fog}  &  \textbf{Rix}  &  \textbf{MLTD} \\
        \midrule
        \noalign{\smallskip} 
        \textsc{AYA23} (Base) & 76.74 &  15.83 &  4.7 &  43.98\\ [3mm]
        \textsc{\textbf{Single Teacher Students}} \\
        \noalign{\smallskip} 
        \textsc{AYA23} & 151.83 & 17.67 & 5.92 &  46.51\\
        \noalign{\smallskip}
        \textsc{LLAMA-3} & 141.71 & 17.33 & 5.87 & 49.5 \\
        \noalign{\smallskip}
        \textsc{GEMMA-2} & 140.59 & 15.67 \tcbox[colback=LightRed]{\textcolor{black}{$\downarrow$}} & 4.28 \tcbox[colback=LightRed]{\textcolor{black}{$\downarrow$}} & 52.48 \\
        \noalign{\smallskip} 
        \textsc{Translation} & 197.05 & 16.62 & 5.22 & \textbf{53.01} \\ [3mm]
        \noalign{\smallskip} 
        \textsc{\textbf{Multilingual Arbitrage}} \\
        \noalign{\smallskip}         
        \textsc{Random Routing} & 144.16 & 17.16 & 5.81 &  45.81\\
        \noalign{\smallskip}
        \textsc{Fixed Routing} & 160.75 & 17.71 & 5.94 &  50.79\\
        \noalign{\smallskip}
        \textsc{Reward Based Routing} & 164.4 & 17.01 & 5.69 &  51.95\\
        \noalign{\smallskip} 
        \textsc{Learned Routing} & \textbf{242.56} & \textbf{19.11} & \textbf{7.74} &  51.08 \\
        \noalign{\smallskip} 
        \noalign{\smallskip}
        \bottomrule
    \end{tabular}}
    \caption{Evaluation of textual characteristics across student models in 4 languages: \textsc{English, German, French and Ukranian}. The number of tokens, Gunning-Fog Index, Rix Index, and Measure of Textual Lexical Diversity (MLTD) for each model highlights the differences in verbosity, readability and lexical diversity. Except for Gemma 2, all students show increase for all metrics.}
    \label{tab:textualcharacteristics-results}
\end{table}

\subsection{Student Model Generations}
\label{app:student_gen}

A list of example model generations in Turkish from different student models can be found in Table \ref{tab:example_student_gen_tur}.
\begin{table}[h]
\centering
\renewcommand{\arraystretch}{1.5}
\scalebox{0.85}{
\begin{tabular}{m{3cm}|m{15cm}}
\toprule
\multicolumn{2}{p{18cm}}{\textbf{Prompt:} \textit{Kedilerin, insanlara kıyasla nispeten az sayıda tat alma tomurcuğu vardır (insan dilindeki 9.000'den fazla ile karşılaştırıldığında yaklaşık 470). Evcil ve yabani kediler tat alıcı gen mutasyonunu paylaşıyor. Bu da tatlı tat alma tomurcuklarının şekerli moleküllere bağlanmasını engelliyor ve tatlı tadı alma yeteneklerini kaybediyor. Onun yerine tat alma tomurcukları asitlere, protein gibi amino asitlere ve acı tatlara tepki verir. Kediler ayrıca yiyecekleri için belirgin bir sıcaklık tercihine sahiptir, taze bir öldürme ile benzer olan yaklaşık 38 °C (100 °F) sıcaklığındaki yiyecekleri tercih eder ve rutin olarak soğuk veya buzdolabında sunulan yiyecekleri reddeder (ki bu, ``av'' öğesinin uzun zamandır ölü olduğunu ve bu nedenle muhtemelen toksik veya çürümeye başladığını kedilere işaret eder). Neden kediler yemek konusunda bu kadar seçicidir?}} \\
\midrule
\noalign{\smallskip}
\textbf{Student Model} & \textbf{Generation} \\
\noalign{\smallskip}
\hline
Single Aya 23 & Kediler, yiyecekleri için belirgin bir sıcaklık tercihine sahiptir ve taze bir avın sıcaklığını tercih ederler. \\ 
\hline
Random Routing & Kediler, yiyecekleri için belirgin bir sıcaklık tercihine sahiptir ve taze bir öldürme ile benzer olan yaklaşık 38 °C (100 °F) sıcaklığındaki yiyecekleri tercih ederler. Bu, kedilerin yiyecekleri için belirli bir sıcaklık aralığına sahip oldukları ve bu sıcaklıkta yiyecekleri tercih ettikleri anlamına gelir. Kediler ayrıca soğuk veya buzdolabında sunulan yiyecekleri reddederler, bu da yiyecekleri için belirli bir sıcaklık aralığına sahip oldukları ve bu sıcaklıkta yiyecekleri tercih ettikleri anlamına gelir. Bu, kedilerin yiyecekleri için belirli bir sıcaklık aralığına sahip oldukları ve bu sıcaklıkta yiyecekleri tercih ettikleri anlamına gelir. \\ 
\noalign{\smallskip}
\hline
Reward-Based Routing & Kediler, yiyecekleri için belirgin bir sıcaklık tercihine sahiptir ve taze bir öldürme ile benzer olan yaklaşık 38 °C (100 °F) sıcaklığındaki yiyecekleri tercih ederler. Bu, kedilerin yiyecekleri için bir sıcaklık tercihine sahip oldukları ve bu sıcaklıkta yiyecekleri tercih ettikleri anlamına gelir. Kediler ayrıca soğuk veya buzdolabında sunulan yiyecekleri reddederler, bu da yiyeceklerin ``av'' öğesinin uzun zamandır ölü olduğunu ve bu nedenle muhtemelen toksik veya çürümeye başladığını kedilere işaret ettiği anlamına gelir. Bu nedenle, kediler yiyecek konusunda bu kadar seçici olabilirler. \\ 
\hline
Learned Routing & Kediler, yiyecekleri için belirgin bir sıcaklık tercihine sahiptir ve taze bir öldürme ile benzer olan yaklaşık 38 °C (100 °F) sıcaklığındaki yiyecekleri tercih ederler. Bu, yiyeceklerin taze ve toksik olmamasını garanti eder. \\
\bottomrule
\end{tabular}}
\caption{Comparison of student model responses trained using different methods: Single Aya 23 (best Turkish teacher), random routing, reward-based routing, and learned routing. The Aya 23 student's answer is too short and inadequate, while random-routing generates responses that are repetitive.}
\label{tab:example_student_gen_tur}
\end{table}

A list of example model generations in English from different student models can be found in Table \ref{tab:example_student_gen}.
\begin{table}[h]
\centering
\renewcommand{\arraystretch}{1.5}
\scalebox{0.85}{
\begin{tabular}{m{3cm}|m{15cm}}
\toprule
\multicolumn{2}{p{18cm}}{\textbf{Prompt:} \textit{Cats have relatively few taste buds compared to humans (470 or so versus more than 9,000 on the human tongue). Domestic and wild cats share a taste receptor gene mutation that keeps their sweet taste buds from binding to sugary molecules, leaving them with no ability to taste sweetness. Their taste buds instead respond to acids, amino acids like protein, and bitter tastes. Cats also have a distinct temperature preference for their food, preferring food with a temperature around 38 °C (100 °F) which is similar to that of a fresh kill and routinely rejecting food presented cold or refrigerated (which would signal to the cat that the ``prey'' item is long dead and therefore possibly toxic or decomposing). Why are cats so picky about what they eat?}} \\
\midrule
\noalign{\smallskip}
\textbf{Student Model} & \textbf{Generation} \\
\noalign{\smallskip}
\hline
Single Llama 3 & Cats are known for being picky eaters, and their unique taste preferences are due to a combination of factors. The main reason is that cats have a specific temperature preference for their food, which is around 38°C (100°F). This temperature is similar to the temperature of a fresh kill, which is around 37°C (98°F) to 40°C (104°F). Cats are also born with a natural instinct to hunt and kill prey, and they are trained to associate certain temperatures with food that is fresh and safe to eat. In addition, cats have a limited number of taste buds compared to humans, which is around 470 taste buds on the tongue. This is because cats have a unique taste receptor gene mutation that prevents their sweet taste buds from binding to sugary molecules. Instead, their taste buds respond to acids, amino acids, and bitter tastes. This means that cats are not as sensitive to sweet tastes as humans are, and they are more likely to be attracted to foods that have a strong flavor or aroma. Another factor that contributes to cats' picky eating habits is their ability to detect and avoid potential toxins. Cats have a natural instinct to avoid food that is toxic or decomposed, and they are trained to associate certain temperatures and textures with potential toxins. This is why cats often reject food that is presented cold or refrigerated, as it may indicate that the food is no longer fresh or safe to eat. Overall, cats' picky eating habits are a result of their unique biological and behavioral characteristics, including their temperature preference, limited number of taste buds, and ability to detect and avoid potential toxins. \\
\hline
Random Routing & They have a distinct temperature preference for their food. \\ 
\noalign{\smallskip}
\hline
Reward-Based Routing & According to the text, cats are picky about what they eat because they have a distinct temperature preference for their food. They prefer food with a temperature around 38 °C (100 °F), which is similar to that of a fresh kill and routinely reject food presented cold or refrigerated. This is because cats have a genetic mutation that prevents their sweet taste buds from binding to sugary molecules, leaving them with no ability to taste sweetness. Their taste buds instead respond to acids, amino acids like protein, and bitter tastes. \\ 
\noalign{\smallskip}
\hline
Learned Routing & Cats are picky eaters due to several factors related to their biological and evolutionary traits: 
\textbf{Taste Sensitivity:} Cats have fewer taste buds than humans and cannot taste sweetness due to a genetic mutation. Their taste buds are more responsive to acids, amino acids, and bitter tastes, aligning with their carnivorous diet.
\textbf{Temperature Preference:} Cats prefer food at around 38°C (100°F), similar to a fresh kill, and often reject cold or refrigerated food, which they may perceive as old and potentially unsafe. These factors ensure that cats consume food that meets their nutritional needs and is safe, reflecting their evolutionary adaptations as obligate carnivores. \\
\bottomrule
\end{tabular}}
\caption{Comparison of student model responses trained using different methods: Single Llama 3 (best English teacher), random routing, reward-based routing, and learned routing. The Single Llama 3 student's answer is too long and includes a lot of irrelevant details, while random routing generates responses that are too brief and insufficient.}
\label{tab:example_student_gen}
\end{table}

\end{document}